\documentclass[5p]{article}
\usepackage[utf8]{inputenc}
\usepackage{amsmath, amsfonts, amsthm}
\usepackage{stmaryrd}
\usepackage{mathtools}
\usepackage{todonotes}
\usepackage{xcolor}
\usepackage{booktabs}
\usepackage{url}
\usepackage[ruled,vlined,linesnumbered]{algorithm2e}

\usepackage[backend=biber,style=numeric]{biblatex}
\newtheorem{definition}{Definition}

\newtheorem{observation}{Observation}
\newtheorem{lemma}{Lemma}
\newtheorem{theorem}{Theorem}
\newtheorem{corollary}{Corollary}

\usepackage{xspace}
\newcommand{\mathsymbol}[2]{\newcommand{#1}{\ensuremath{\mathit{#2}}\xspace}}
\newcommand{\mathmacro}[2]{\newcommand{#1}[1]{\ensuremath{\mathit{#2}}\xspace}}

\mathsymbol{\probOne}{\text{\sc{acjs1}}}
\mathsymbol{\probMulti}{\text{\sc{acjsm}}}
\newcommand{\prob}[3]{\ensuremath{#1| #2| #3}}
\newcommand{\probWC}[2]{\prob{#1}{#2}{\sum w_jC_j}}
\newcommand{\probWF}[2]{\prob{#1}{#2}{\sum w_jF_j}}
\mathsymbol{\probGraham}{\probWF{P_m}{r_j\leq d}}
\mathsymbol{\probGrahamNoDeadlines}{\probWC{P_m}{}}

\mathsymbol{\jobsPriority}{J^-}
\mathsymbol{\jobsLate}{J^+}
\mathmacro{\jobsPriorityWithIndex}{J^-_{#1}}
\mathmacro{\jobsLateWithIndex}{J^+_{#1}}
\mathsymbol{\jobsPriorityWithoutGiant}{\jobsPriorityWithIndex{-g}}
\mathsymbol{\jobsLateWithoutGiant}{\jobsLateWithIndex{-g}}

\title{Single and Parallel Machine Scheduling with Variable Release Dates}
\author{Felix Mohr, Gonzalo Mejía, Francisco Yuraszeck}
\date{March 2020}

\begin{document}


\maketitle

\begin{abstract}
In this paper we study a simple extension of the total weighted flowtime minimization problem for single and identical parallel machines.
While the standard problem simply defines a set of jobs with their processing times and weights and assumes that all jobs have release date 0 and have no deadline, we assume that the release date of each job is a decision variable that is only constrained by a single global latest arrival deadline.
To our knowledge, this simple yet practically highly relevant extension has never been studied.
Our main contribution is that we show the NP-completeness of the problem even for the single machine case and provide an exhaustive empirical study of different typical approaches including genetic algorithms, tree search, and constraint programming.
\end{abstract}

\section{Introduction}
In this paper, we address the identical parallel machine scheduling problem with variable release dates and a common deadline for arrival.
This problem occurs in several settings in which the release dates themselves are decision variables with the constraint that all jobs must arrive before or on a common fixed deadline. This deadline can be interpreted as a maximum release date for all jobs.

To our knowledge, this problem has not been studied before in spite of many important applications.
A first example is a manufacturing facility which uses a Just-In-Time discipline: jobs are released to the shop floor as late as possible to avoid cluttering the system but due to accounting restrictions, mostly related to the MRP (Materials Requirements Planning) logic, all work orders in a time bucket must be released before a fixed deadline.
A second example is the receiving area of a warehouse which restricts the arrival of trucks within a time window.
The warehouse may schedule its suppliers' trucks so to avoid congestion and provide them with an arrival time, but again, the warehouse's opening hours or external constraints such as circulation bans at certain hours, restrict the arrival of trucks.
In these two examples, the deadline constraint cannot be violated, and a central controller must guarantee that all jobs meet such a constraint.

Our contribution is threefold:
\begin{itemize}
    \item We show that the studied problem is NP-complete even for the one machine case.
    \item We analyze the problem empirically comparing scores of optimal schedules to optimal scores of the same instances in the standard problem (without deadlines).
    
    \item We devise problem formulations for typically adopted search algorithms including neighborhood-based search algorithms, constraint programming, and tree search and conduct a broad experimental comparison.
\end{itemize}

\section{Problem Statement}
\label{sec:problem}
The problem is defined as follows.
Let $J$ be a set of jobs indexed in $j$, each with a processing time $p_j$ and positive weight $w_j$.
The release date $r_j$ of each job can be set (is a decision variable) but cannot be greater than a common arrival deadline $d \geq 0$.
Every job must be allocated on exactly one machine of a given set $I$ of machines indexed in $i$.
The objective is to minimize the total weighted flowtime $\sum w_jF_j$, where $F_j$ is defined as the completion time of job $C_j$  minus its release date $r_j$.
In the notation of Graham et al. \cite{graham1979optimization}, we contrast our problem to \probWF{P_m}{r_j}, in which  $r_j$ is a parameter, by making the deadline $d$ explicit and hence emphasizing that $r_j$ are \emph{decision variables}, yielding the notation \probGraham.

\begin{figure}
    \centering
    \includegraphics[width=\columnwidth, trim=0 14 0 0, clip]{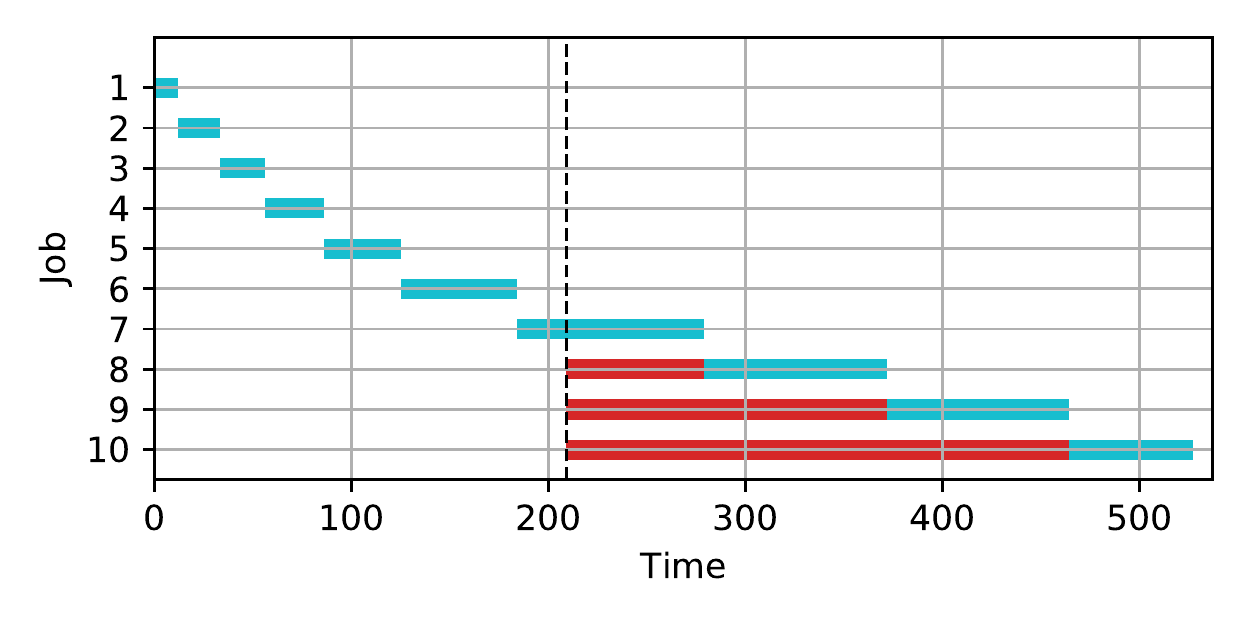}
    \caption{Sample Schedule for 10 jobs. Clearly, the jobs starting earlier that the deadline $d$ (dashed line) arrive just in time while the other jobs arrive \emph{exactly} at time $d$ and wait until they are processed.}
    \label{fig:sampleschedule}
\end{figure}

Interestingly, even though the release dates of the jobs are decision variables, we can in fact see them rather as an implicit consequence of the only binary the decision of whether a job should be processed prior to $d$ or not.
The reason is that we will of course choose $r_j$ as close to the start time as possible, which implies for jobs starting prior to $d$ that they arrive just in time, and for jobs starting not prior to $d$ that they arrive \emph{exactly} at $d$.
Fig. \ref{fig:sampleschedule} depicts this rationale for a sample schedule in our problem:
Jobs starting prior to $d$ arrive just in time, while the others arrive exactly at time $d$ and remain in the "buffer" (red part) until being processed.

Based on this observation, we start the mathematical problem formulation by first considering the cost induced by a schedule $\pi$ on a single machine:
\begin{equation}
    \label{eq:objectivefunction}
    \phi(\pi) = \left(\sum\limits_{j \in \jobsPriority} w_jp_j\right) + \smashoperator{\sum\limits_{j \in \jobsLate}} w_j\left(
    \smashoperator[r]{\sum_{k \leq_\pi j}}p_k - d\right)
\end{equation}


For any schedule $\pi$, the set of jobs starting strictly prior to the deadline $d$ is uniquely defined.
Here, and in the rest of the paper, we denote this set as $\jobsPriority \subseteq J$.
The complementary set of jobs starting not earlier than the deadline is denoted as $\jobsLate = J \setminus \jobsPriority$.
Even though \jobsLate and \jobsPriority depend on a concrete schedule $\pi$, we omit this schedule in the notation since it is always clear from the context.

The interpretation of this cost function is as follows.
The first term takes into account that all jobs in \jobsPriority arrive just-in-time (i.e. release date = start time) and hence their contribution to the overall cost is only their own weighted processing time.
The second summand contains the calculation of completion times of jobs in \jobsLate.
Each of those jobs arrives \emph{exactly} at time $d$ as there is no reason to arrive earlier and they are not allowed to arrive later.
Hence, the flowtime of job $j \in \jobsLate$ is simply the sum of all jobs processed before ($k <_\pi j$) and $j$ itself minus the deadline (and hence arrival time of job $j$) $d$.

We then define as \probGraham the optimization problem of finding
\begin{equation}
    \min\limits_{\pi_1,..,\pi_m}~~~~ \sum\limits_{i=1}^m \phi(\pi_i)
\end{equation}
where $\dot\bigcup_{i=1}^m \pi_i = J$.

That is, we can understand each schedule $\pi_i$ as an ordered set of jobs, and each job must be contained in exactly one schedule.
In the rest of the paper, we will commonly treat schedules $\pi_i$ as ordered sets and hence apply set operations such as unions on them.


Observe that our problem is very similar to the total weighted \emph{tardiness} minimization problem.
To see this, please observe that the second summand in target function in Eq. (\ref{eq:objectivefunction}) in fact \emph{is} the total weighted tardiness for all jobs in \jobsLate, if we set a common due date $d$ for them.
Now the difference is that, in our case, the jobs starting prior to the $d$ do contribute with their weighted processing time, regardless of whether they finish prior to $d$ or not or by the difference between their completion time $C_j$ and $d$.

In spite of their similarity, which might even indicate isomorphism between the two problems, a simple example shows that the problems are in fact distinct.
Consider a single machine problem  processing two jobs with the parameters $w_1 = 10, p_1 = 2, w_2 = p_2 = 1$ and a deadline $d = 1$. Let  $\pi_1 = (1,2)$ and $\pi_2 = (2,1)$ be the two possible schedules.
It can be noticed that the objective function values of $\pi_1$ and $\pi_2$ are $12$ and $20$ respectively, so $\pi_1$ is the better schedule.
On the contrary, the metric in Eq. (\ref{eq:objectivefunction}) assigns $22$ to $\pi_1$ and $21$ to $\pi_2$, so $\pi_2$ is the better choice.

\section{Related Work}
\label{sec:relatedwork}
Scheduling problems involving parallel machines have been studied for decades now and there is a vast body of literature on the topic.
The literature distinguishes identical ($P_m$), uniform ($Q_m$), and unrelated ($R_m$) parallel machines.
The first surveys date from the early nineties \cite{Cheng1990}. 
The reader is referred to the comprehensive works of \cite{Brucker2005, doi:10.1287/moor.22.3.513, Elmaghraby1974, Jouglet2011, mokotoff2004exact} for more information.

Regarding the complexity of problems involving total completion or flow time minimization, it is well known that several single or parallel machine scheduling problems can be solved optimally in polynomial time. For example, the Weighted Shortest Processing Time rule (WSPT) solves the classical \probWC{1}{} and the Shortest Processing Time rule (SPT) rule in combination with a list scheduling algorithm solves the \prob{P_m}{}{\sum C_j}.
A special case solved in pseudo-polynomial time is the \probWC{P_m}{p_j=1} \cite{Baptiste2008}.
However, the vast majority of problems that are built upon identical parallel machines such as \probWC{P_m}{} \cite{DBLP:journals/cacm/BrunoCS74}, \probWC{P_m}{s_{kj}} \cite{Baez2019, Kramer2019a}, \probWC{P_m}{batch} \cite{Jia2018, Arroyo2019, Zhang2020} and  \prob{P_m}{}{\sum w_j(E_j+T_j)} \cite{Cheng1994} are NP-hard.
Notice that we used the standard notation of $s_{jk}$, $T_j$ and $E_j$ respectively for sequence dependent setup times, tardiness and earliness.

Most works on identical and uniform parallel machines and weighted flowtime minimization (i.e. \probWC{P_m}{} and \probWC{Q_m}{}) examine exact algorithms.
Most notably, \probWC{P_m}{} has been attacked successfully by deriving strong lower bounds and using Branch and Bound (B\&B) \cite{Elmaghraby1974, sarin1988improved, azizoglu99}.
Other methods for parallel machine scheduling problem consider Branch and Price (B\&P) algorithms \cite{chen1999solving}.

While most of these fundamental works date back to the last century, the case of \probWF{P_m}{r_j}, i.e., with (parametric) release dates, has been studied more recently over the last two decades.
\cite{Yalaoui2006}   investigated dominance rules and developed a B\&B algorithm able to solve relatively large instances. Later \cite{Baptiste2008} proposed new lower bounds and \cite{Nessah2008} showed an algorithm able to handle instances up to 100 jobs and two machines.
A related work is that of \cite{Lee2016} who also proposed a B\&B algorithm for competing agents (i.e. when jobs from different entities have different priorities). 
A more recent paper of \cite{Kramer2020} addressed the \probWF{P_m}{r_j} with a B\&P algorithm and solved to optimality instances of up to 200 jobs and 10 machines with small optimality gaps on larger instances.
\probWF{P_m}{r_j} has also been investigated through the notion of pricing by \cite{doi:10.1287/opre.47.6.862} and by  \cite{Kowalczyk2018}.
The latter authors proposed new branching schemes and suppression mechanisms for the pricing sub-problem.
The generic work of \cite{bulhoes2020exact} studied a larger class of parallel machines with identical, uniform and unrelated machines with regular and non-regular objectives. They proposed a B\&P algorithm with new cuts and proved optimality (previously unknown) for a number of classical instances.

Other approaches for identical and non-identical parallel machine scheduling problems are mathematical formulations: \cite{Unlu2010} evaluated different Mixed Integer Linear Programming (MILP) with different variable types (completion times, time indexed, linear ordering and network variables) and with different objective functions that include total weighted completion time.
Later, \cite{ahmed2013scheduling} proposed another math formulation for \probWF{P_m}{ r_j}  and a heuristic algorithm. 
Recent works \cite{Bulbul2017, FANJULPEYRO2019173}  developed new formulations for the unrelated case (which it is clearly a special case of the identical case) and arc-flow formulations \cite{Kramer2019} for the \probWF{P_m}{r_j} case.
Recently, \cite{malapert2019new, Nattaf2019} examined Constraint Programming methods for parallel machines scheduling problems although not for the particular case of \probWF{P_m}{r_j}.
Other researchers (e.g.\cite{Lee2014, Rodriguez2012, Vallada2011, Zaidi2010}) have proposed heuristics and meta-heuristics for scheduling problems with unrelated parallel machines and total flowtime minimization

To our knowledge, the work on problems with flexible or variable release dates is quite limited.
One work that explicitly treats release dates as flexible is \cite{Nurre2018}.
Even though not introduced in this manner, similar to our paper, release dates are also considered implicit decision variables that are derived from other decisions.
However, being an approach applied for network optimization problems, the role of release dates in \cite{Nurre2018} is different in that their values are limited by maintenance constraints of the network structure.
Similarly, \cite{elvikis2007scheduling} consider the release dates as dependent of location decisions.
In both cases, the considered metric is different from total weighted flowtime.
Apart from these, we are not aware of any work that treats release times as variables.

\section{Theoretical and Empirical Problem Analysis}
In this section, we analyze the studied problem in depth.
First, in Sec. \ref{sec:analysis:npcompleteness}, we show the NP-completeness.
Building upon this, Sec. \ref{sec:analysis:theoreticalinsights} derives further theoretical results and discusses the difficulty of creating lower bounds for this problem as opposed to the standard problem.
Finally, Sec. \ref{sec:analysis:empirical} provides empirical insights into the problem by comparing its optimal solutions with solutions that would be optimal in the standard problem, and also shade light on the role of the deadline $d$ for this relation.

\subsection{NP-Completeness of the Problem}
\label{sec:analysis:npcompleteness}
For the single machine case, the stated problem is a combination of two simpler problems that becomes non-trivial because of the deadline $d$.
These two problems stem from the sets of instances defined by two extreme values of $d$.
First, the set of instances with $d = 0$ constitutes \probWC{1}{}, which is solved optimally using the WSPT (Weighted Shortest Processing Time) rule in $\mathcal{O}(n  \log n)$ \cite{smith1956various}.
On the other hand, if $d \geq \sum_{j \in J} p_j$, we will have $\jobsPriority = J$ and hence $\phi(\pi) = \sum_{j\in J}w_j p_j$ for \emph{all} schedules $\pi$, which is the absolute minimum possible.
However, for other $d$ in between, it is not so obvious how to proceed.

We will take the following insight for granted, generally excluding solutions with idle times from the discussion.
\begin{observation}
    \label{theorem:noidletimes}
    Every instance of \probGraham has an optimal solution that has no inserted idle time on any machine.
\end{observation}

\noindent
A second insight that will be important is the following:

\begin{lemma}
\label{theorem:swpt}
In any optimal solution to \probGraham, the jobs starting not strictly earlier than the deadline are sorted by the WSPT rule on each machine.
\end{lemma}
\begin{proof}
    The problem of ordering the jobs that start after the deadline is simply a new scheduling problem in which the machine becomes available only after some time $l$ (previously it is busy with the last job that started before or on the deadline).
    By linearity, the length of $l$ does not matter and we can set it w.l.o.g. to 0.
    But then we have created a standard problem without deadline with a trivial optimal solution given by WSPT.
\end{proof}

To analyze the complexity of \probGraham, we first define its decision problem variant, which we call the \emph{arrival-constrained job scheduling problem} ({\sc acjsp}).
\begin{definition}
    Define as
    
    \begin{itemize}
        \item \probOne the question whether, given a problem instance as above with $m=1$ and an upper bound $y \in \mathbb{R}$, there exist a schedule $\pi$ such that $\phi(\pi) \leq y$, and as
        \item \probMulti the question whether, given a problem instance as above with $m\geq 2$ and a threshold $y \in \mathbb{R}$, there exist schedules $\pi_1,..,\pi_m$ such that $\dot\cup_{j=1}^m \pi_j = J$ and $\sum_{j=1}^m\phi(\pi_j) \leq y$.
    \end{itemize}

\end{definition}


We will show the NP-completeness of \probOne and \probMulti as follows.
First, we will derive the NP-completeness of \probOne by a reduction from the subset sum problem.
The subset sum problem consists of deciding whether for a given set of positive integers $a_1,..,a_n$ and an integer $b$ there exists a subset $S \subseteq \{1,..,n\}$ such that $\sum_{i\in S} a_i = b$, and it is known to be NP-complete \cite{karp1972reducibility}.
The NP-completeness of \probMulti is then a trivial consequent of this result.

The main difficulty in this reduction is to establish an upper bound $y$ for the score function $\phi$ that can be achieved if and only if there exists a decomposition such that $\sum_{i \in S} a_i = b$.
The problem is that, in contrast to other reductions to similar problems \cite{lenstra1977complexity}, our score function $\phi$ is heterogeneous in that it scores the jobs starting prior to the deadline in a substantially different way than the jobs starting after it.
We conjecture that, opposed to many other reductions in the area of scheduling, it is not always possible to provide a \emph{tight} value for $y$ in the sense that $y$ is the \emph{true best} solution if a decomposition of $a_i$ to $b$ exists and not reachable if it does not exist.

The solution to this problem lies in the adoption of an auxiliary \emph{giant} job.
Such a giant job is so huge in both processing time and weight that it essentially degenerates the problem instance containing it to the question of whether there exists a selection $S$ of the other (normal) jobs such that the sum of processing times of jobs in $S$ is \emph{exactly} the deadline $d$.
The reduction will then consist of deriving an appropriate giant job for a given instance of the subset sum problem.
The presence of a giant job will allow us to determine a value of $y$ that can be achieved iff $\sum_{i \in S} a_i = b$.

\begin{definition}
    \label{def:giant}
    An instance of \probOne in which $p_j, w_j \geq 1$ for all $j \in J$ is said to contain a \emph{giant} job if there is a distinguished job $g\in J$ such that
    \begin{enumerate}
        \item $p_g > 2 \smashoperator{\sum\limits_{j,k \in J\setminus\{g\}}}w_jp_k$ and 
        \item $w_g > p_g \max\limits_{j \in J \setminus\{g\}} w_j$.
    \end{enumerate}
    In such a setting, we denote as $J_{-g} = J \setminus \{g\}$ the set of jobs without the giant.
    Accordingly, we will denote \jobsPriorityWithoutGiant and \jobsLateWithoutGiant as the respective job sets without the giant.
\end{definition}
Intuitively, a giant job has a processing time that is larger than the dot product of weights and processing times of all jobs excluding the giant job, and its weight exceeds its own processing time multiplied by the largest of the other jobs' weights.
Note that, in our reduction from subset sum, we will always have $w_j,p_j \geq 1$ for all jobs $j \in J$ as required in the definition of a giant job.

\begin{observation}
\label{theorem:giantobservations}
For every instance of \probOne with a giant job,
\begin{enumerate}
    \item the giant job is unique
    \item $w_g$ and $p_g$ are not related to the deadline $d$
    \item $d < \sum_{j\in J_{-g}} p_j$; otherwise the problem has a trivial optimal solution.
    \item \label{theorem:giantobservations:wspt} the WSPT rule orders the giant job prior to any other job
\end{enumerate}
\end{observation}

These properties are obvious and we omit a proof.

\begin{lemma}
\label{theorem:giantjobnotlaterthandeadline}
A schedule $\pi$ for an instance of \probOne containing a giant job $g \in J$ is sub-optimal if $g$ is started strictly later than the deadline $d$.
\end{lemma}
\begin{proof}
    Assume that $g$ starts strictly later than the deadline $d$. Then it is the \emph{first} job to start later than $d$; otherwise $\pi$ is sub-optimal by Lemma \ref{theorem:swpt} and the observation that $g$ is first scheduled by the WSPT rule (point (\ref{theorem:giantobservations:wspt}) in Observation \ref{theorem:giantobservations}).
    This implies that there must be another job $j$ directly preceding $g$ and starting not later than $d$.
    Consider then the schedule $\pi'$ that is equal to $\pi$ only that $j$ and $g$ are swapped.
    We observe that
    \begin{align*}
        \phi(\pi) & \geq \sum\limits_{k\in \jobsPriority} w_kp_k + w_jp_j  + w_g(1+p_g) + M\\
        \phi(\pi') & \leq \sum\limits_{k\in \jobsPriority} w_kp_k + w_gp_g + w_j(p_j+p_g) + M,
    \end{align*}
    where the 1 in the first inequality results from the fact that the giant job $g$ starts strictly later than $d$, so the factor multiplied to its weight is at least its own processing time plus 1.
    The term $M$ refers to the contribution of the jobs coming after $g$, and their contributions are not affected by the swap.
    
    Since the leading sums and $M$ cancel out, we get that
    \begin{align*}
        \phi(\pi') - \phi(\pi) &\leq w_gp_g + w_j(p_j+p_g) - (w_jp_w + w_g(1+p_g))\\
        & = w_jp_g - w_g < 0,
    \end{align*}
    where the last inequality follows directly from property (2) of the definition of the giant job (Def. \ref{def:giant}).
\end{proof}

From this, we obtain the following result:
\begin{lemma}
    \label{theorem:optimalfit}
    If an instance of the \probOne contains a giant job $g$, 
    then there exists an optimal schedule $\pi^*$ for the instance such that $\phi(\pi^*) \leq w_gp_g + \smashoperator{\sum\limits_{j,k\in J_{-g}}} w_kp_j + \smashoperator{\sum\limits_{k \in \jobsLateWithoutGiant}} p_gw_k$.
\end{lemma}
\begin{proof}
    By Lemma \ref{theorem:giantjobnotlaterthandeadline}, the giant jobs starts not later than the deadline, so its weight $w_g$ contributes only for its own processing time $p_g$.
    The second term is simply a conservative bound for the total weighted flowtime of all jobs ignoring the influence of the giant job.
     The last term accounts for the fact that all jobs starting after the deadline (the set \jobsLate) are delayed in their start by the degree to which the giant job reaches into the slot after the deadline.
     This is at most $p_g$, which occurs when the giant job starts exactly at the deadline.
\end{proof}

\begin{theorem}
    \label{theorem:npcompleteness}
    \probOne is NP-complete. 
\end{theorem}
\begin{proof}
    
    \emph{NP membership.}
    Clear since a solution can be guessed in polynomial time, and its evaluation just requires a call of $\phi$, which is polynomial in the problem size.
    
    \emph{NP-Hardness.}
    By reduction from the subset-sum problem, which is known to be NP-complete \cite{karp1972reducibility}.
    Given $a_1$,..,$a_n$, $b \in \mathbb{N}$, let $A = \sum_{i=1}^n a_i$, and we encode the \probOne instance as follows:
    \begin{itemize}
        \item[ ] $J = \{1,..,n,n+1\}$. Fix $g = n+1$ to refer to the giant job
        \item[ ] $d = b$
        \item[ ] $p_i = w_i = a_i$ for $1 \leq i \leq n$
        \item[ ] $p_g = 2 A^2 + 1$,\hspace{1em}
        $w_g = p_g \max\limits_{i\in\{1,..,n\}} a_i + 1$.
        \item[ ] $y = A^2 + p_g(A-b + w_g)$
    \end{itemize}
    The job $g = n+1$ is a giant job by construction.
    
    If there \emph{does} exist an $S \subseteq \{1,..,n\}$ such that $\sum_{i\in S}a_i = b$, then there is a schedule $\pi$ with $\sum_{i\in \jobsLateWithoutGiant} w_i = \sum_{i\in \jobsLateWithoutGiant} a_i = A - b$.
    Plugging this into the definition of $y$, Lemma \ref{theorem:optimalfit} yields that there is a $\pi$ such that $\phi(\pi) \leq y$.
    
    Now consider the case that such a decomposition does \emph{not} exist.
    Then the giant job can in no optimal schedule start exactly at time $d$ (cf. Observation \ref{theorem:noidletimes}), which implies by Lemma \ref{theorem:giantjobnotlaterthandeadline} that it must start \emph{strictly} earlier.
    From this, however, we can infer that $\sum_{i\in     \jobsPriorityWithoutGiant} a_i < b$ for any schedule and hence $\sum_{i\in \jobsLateWithoutGiant} a_i > A - b \Rightarrow \sum_{i\in \jobsLateWithoutGiant} a_i  \geq A - b + 1$.
    But then we have for \emph{any} schedule     $\pi$ that
    \begin{align*}
        \phi(\pi) & > w_gp_g  + (p_g -d) \smashoperator{\sum\limits_{j\in  \jobsLateWithoutGiant}} a_j\\
        & > w_gp_g  + (p_g -d)(A - b + 1)\\
        & = p_g (A - b + w_g) + p_g - d(A-b+1)\\
        & = p_g (A - b + w_g) + 2A^2 + 1 - b(A-b+1)\\
        & = y + A^2 + 1 - b(A - b +1)\\
        & \geq y + A^2 + 1 - A(A - b + 1)\\
        & = y + 1 +A(b - 1) \geq y + 1 > y
    \end{align*}
\end{proof}

We can now directly derive the NP-completeness for the case of several machines.

\begin{corollary}
    \probMulti is NP-complete.    
\end{corollary}
\begin{proof}
    NP membership is still clear, because guessing a solution is also here possible in polynomial time, and checking such a solution only requires linear time by evaluating $\phi$ for a number of $m$ times. The NP-hardness follows from a reduction from \probOne setting $m = 1$.
\end{proof}
Note that the NP-completeness for \probMulti can also directly be established by reducing from the standard problem of parallel machines, which is also NP-complete \cite{lenstra1977complexity}, by setting $d = 0$.

\subsection{Theoretical Insights for the Solution Construction}
\label{sec:analysis:theoreticalinsights}
This section is above all dedicated to the role of the WSPT rule for the construction of optimal solutions.
Building upon the results of the previous section, we show how the NP-completeness of the single machine case has important implications in the finding of strategies to attack the problem that make use of the WSPT rule.
We give examples of curious sub-optimality of WSPT but also highlight its potential importance for solution approaches.

Prior to further analysis, we give two examples of sub-optimality of the WSPT rule for the single machine case.
To this end, consider the two problems shown in the left and right table of Fig. \ref{fig:wsptsuboptimality}.
The first problem is for a deadline of 9, and the description is already sorted by WSPT: the schedule $(0, 1, 2, 3)$ has a score of $78$.
However, it can be easily seen that \emph{all} jobs can start prior to the deadline, e.g. in $(0,2,3,1)$, yielding the score is $76$ consisting only of the sum of weighted processing times of the jobs.
The second problem, which is for deadline 120 and also already sorts the jobs via WSPT, we do not have such a trivial solution.
However, the score of the WSPT solution $(0,1,2,3,4)$ is $17969$, whereas it is $15980$ for the schedule $(2,4,1,0,3)$; a cost reduction of over 11\%.

\begin{figure}[t]
    \centering
    \begin{tabular}{c|cccc}
     j & 0 & 1 & 2 & 3\\\hline
     p & 3 & 6 & 2 & 3\\\hline
     w & 5 & 9 & 2 & 1
    \end{tabular}
    \hspace{1cm}
    \begin{tabular}{c|ccccc}
     j & 0 & 1 & 2 & 3 & 4\\\hline
     p & 18 & 37 & 16 & 88 & 49\\\hline
     w & 63 & 95 & 24 & 96 & 51
    \end{tabular}
    \caption{Example problems with sub-optimal WSPT schedules.}
    \label{fig:wsptsuboptimality}
\end{figure}

Now the first conclusion we can derive from the result of the previous section is that we cannot hope to efficiently \emph{derive} the optimal solution from the WSPT solution.
That is, there cannot be an efficient way of ``repairing'' the WSPT solution by doing a re-ordering of the jobs using only a simple rule:

\begin{corollary}
    \label{theorem:norepairpossible}
    Unless P = NP, there is no efficient way of repairing a WSPT solution for \probOne to an optimal solution.
\end{corollary}
\begin{proof}
    If we had such a repair mechanism, we could combine it with WPST to find an efficient solution to an NP-complete problem and would have shown that P = NP.
\end{proof}

However, the above does certainly not mean that WSPT is entirely useless for our problem.
In fact, the WSPT solution can be a good proxy or even surely optimal in some situations.
One example are problems with the following condition:

\begin{theorem}
    For single-machine problems in which all jobs have the same weight, the schedule produced by WSPT is optimal.
\end{theorem}

\begin{proof}
W.l.o.g. assume $w_j = 1$ for all jobs.
WSPT simply sorts the jobs by their processing time.
Consider a schedule $\pi$ in which $p_j > p_k$, but $j <_\pi k$, and suppose that w.l.o.g. that $k \in \jobsLate$ (otherwise optimality is not affected) and $j$ is the job \emph{directly} preceding $k$.
Moreover, suppose that $j \in \jobsPriority$; otherwise $\pi$ is sub-optimal by Lemma \ref{theorem:swpt}.
Then switching jobs $j$ and $k$ in an otherwise equal schedule $\pi'$ yields a performance difference of $\phi(\pi) - \phi(\pi') = p_j - p_k > 0$; hence $\pi$ was sub-optimal.
\end{proof}

Also for the general case, we will show in the next section that the WSPT solution is often very close to the optimal solution.
Hence, as a seed solution for local searches or a strong upper bound, it plays an important role for the development of other strategies.

With respect to the \emph{methodology} of how to produce solutions in the single-machine case, the NP-completeness and Lemma \ref{theorem:swpt} yield that we essentially need to fix not a full schedule but ``only'' a subset of the jobs to be started prior to the deadline:
\begin{corollary}
    \label{theorem:yieldspartitionproblem}
    For any schedule $\pi$, if $\pi$ is not optimal, then it differs from every optimal schedule in the set \jobsPriority of jobs starting prior to the deadline.
\end{corollary}
\begin{proof}
  The order of jobs starting prior to the deadline has no effect on the performance as long as the re-ordering does not make that one or more jobs now start after the deadline.
  Also, by Lemma \ref{theorem:swpt}, the order of jobs starting \emph{after} the deadline is fixed by WSPT.
  Consequently, a canonical solution in the single-machine case is fully characterized by the set of jobs starting prior to the deadline.
  Hence, if such a solution is not optimal, every optimal solution must have a different such set.
\end{proof}

Together with the NP-completeness of the problem, this implies that orders do not play any role for solving the single machine case but only the subset of jobs starting prior to the deadline.
Of course, one could have attempted to solve our problem as this kind of \emph{partitioning} problem from the very beginning.
However, it is only by the NP-completeness of the problem that we know that there \emph{is no substantially better} way of solving the problem.

Hence, every search algorithm can be designed towards deciding on the partition $\jobsPriorityWithIndex{1} \dot \cup \jobsLateWithIndex{1} .. \cup \jobsPriorityWithIndex{m} \dot \cup \jobsLateWithIndex{m}$.
Here, \jobsPriorityWithIndex{i} and \jobsLateWithIndex{i} refer to the jobs assigned to machine $i$ prior and after the deadline respectively.
Clearly, for each such partitioning there is a canonical and efficiently computable best solution in the space of schedules that satisfy this partitioning: The jobs arriving prior to the deadline (\jobsPriority) can w.l.o.g. be sorted by their processing time; jobs arriving on or after the deadline can be sorted according to WSPT rule yielding a total complexity of $\mathcal{O}(n\log n)$ for the derivation of the best solution for a particular partition.

With respect to the case of identical parallel machines, the NP-completeness of the single-machine problem affects the ability to compute \emph{lower bounds} as required in branch-and-bound solutions.
As discussed in Sec. \ref{sec:relatedwork}, a typical approach to the problem without deadlines is branch and bound, and lower bounds have been established already decades ago \cite{eastman1964bounds,sarin1988improved}.
However, not only can we not \emph{use} these lower bounds on optimal solutions found but also we cannot \emph{derive modified} versions thereof.
The reason is that we know important properties of optimal solutions of the standard problem for parallel machines, but these properties do not hold anymore in \probGraham.
For example, we know that the optimal solution in \probGrahamNoDeadlines still defines a partial order that is compatible with the total ordering defined by WSPT \cite{eastman1964bounds}, and this property is crucial for the typical lower bounds derived in these settings \cite{sarin1988improved}.
However, due to the NP-completeness of the 1-machine case \probOne, this kind of property is not preserved, and there is no way of modifying the approaches for the bounds in the standard case towards \probMulti.
In this sense, even though both problems are NP-complete, we can think of  \probGraham being more difficult than the standard problem \probGrahamNoDeadlines.

At this stage, the best lower bound we have found for the one machine case is the following:
\begin{theorem}
    For every schedule $\pi$ it holds that 
    \begin{equation}
        \phi(\pi) \geq \sum_{j\in J} w_jp_j + \min_j p_j \left(\sum_{j=1}^b (j-1)w_{\sigma(j)} \right),
    \end{equation} where $b$ is a lower bound on $|\jobsLate|$ and $\sigma$ sorts the jobs in $J$ in ascending order of their weights.
\end{theorem}
Note that $b$ can be computed in $\mathcal{O}(n \log n)$ by sorting the jobs by their processing time and checking how many can fit at most into \jobsPriority.

\begin{proof}
\footnotesize
  \begin{align*}
      \phi(\pi) & = \left(\sum\limits_{j \in \jobsPriority} w_jp_j\right) + \smashoperator{\sum\limits_{j \in \jobsLate}} w_j\left(
    \smashoperator[r]{\sum_{k \leq_\pi j}}p_k - d\right)\\
    & = \sum\limits_{j \in J} w_jp_j + \sum\limits_{j \in \jobsLate}w_j\left(\sum_{k <_\pi j}p_k - d\right)\\
    & \geq \sum\limits_{j \in J} w_jp_j + \sum\limits_{j \in \jobsLate}w_j\left(\sum_{k <_\pi j, k \in \jobsLate}p_k\right)\\
    & \geq \sum\limits_{j \in J} w_jp_j + \min_j p_j \sum\limits_{j \in \jobsLate}w_j\left(\sum_{k <_\pi j, k \in \jobsLate}1\right)\\
    & \geq \sum\limits_{j \in J} w_jp_j + \min_j p_j \left(\sum_{j=1}^b w_{\sigma(j)} \left(\sum_{k = 1}^{j-1} 1\right)\right)\\
    & = \sum\limits_{j \in J} w_jp_j + \min_j p_j \left(\sum_{j=1}^b (j-1)w_{\sigma(j)}\right)
  \end{align*}
\end{proof}
This lower bound on any optimal solution is substantially better than any other bound we could observe during our analysis.
Obviously, it is better than the trivial bound of $\sum_{j \in J} w_jp_j$.
Besides, it is also better than the lower bounds resulting by a relaxation of the MILP or bounds derived by the CPSolver during the runs.

However, our empirical analysis found that it is not good enough for practical considerations.
In an experiment series of 1000 instances with 15 jobs on a single machine, we found that the lower bound assumes a value of under 70\% of the optimum in more than 75\% of the cases and even a value of under 60\% in every second case.
These huge gaps do neither constitute a meaningful basis for a B\&B procedure nor for qualifying the algorithm performances.

\subsection{Experimental Problem Analysis}
\label{sec:analysis:empirical}
The fact the WSPT does not deliver optimal results does not mean that it would not return \emph{good} solutions.
In fact, the solutions of WSPT are often even optimal for the one machine case, and we shall empirically show that, at least for rather small problems, the gap between the WSPT solution and the optimal solution is usually small, i.e. on the order of 1\% on average.
\begin{figure*}[t]
    \centering
    \includegraphics[width=\textwidth]{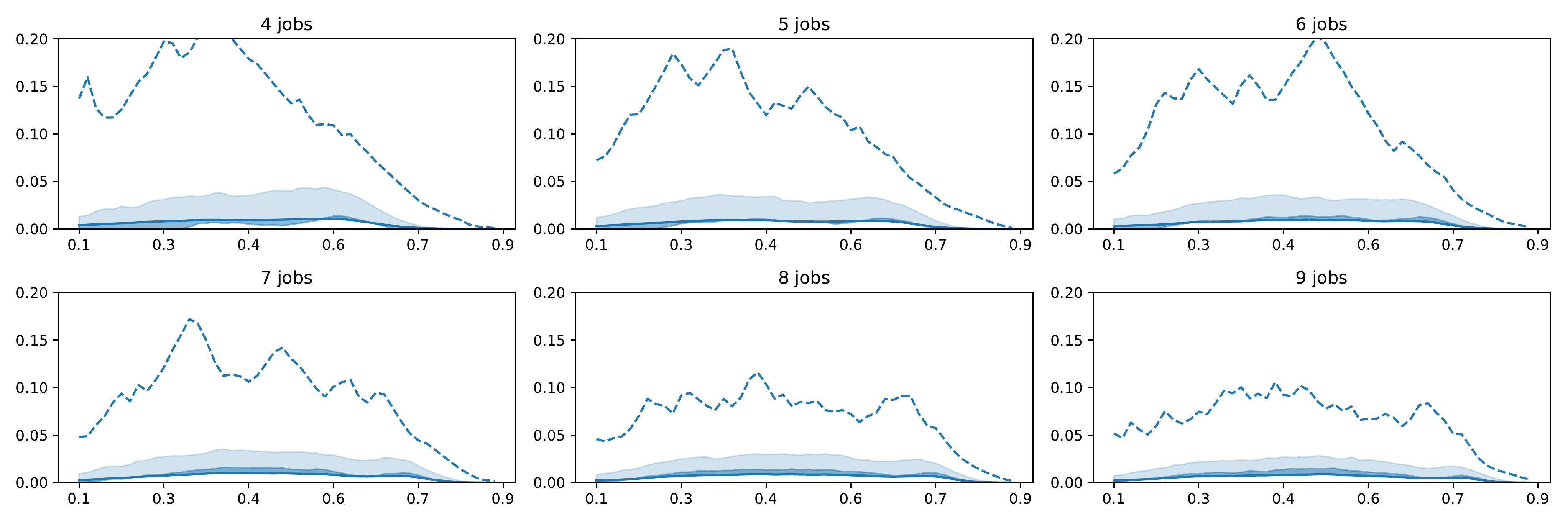}
    \caption{Statistics of the gap between WSPT to and the optimum in the single machine case depending on the relative deadline $d$ (relative to the total processing time of all jobs and hence in $[0,1]$) for 1000 samples.
    The solid lines show the mean gap of the WSPT solution to the optimum solution. The light shaded area shows the .9 quantile, and the dark shaded area the .75 quantile. The dashed line is the maximum gap observed.
    Processing times and weights are sampled uniformly from $\{1,..,100\}$ respectively.}
    \label{fig:wsptgaps}
\end{figure*}

To see this in more depth, consider the plots in Fig. \ref{fig:wsptgaps}.
For single-machine problems with $4$ to $9$ jobs, i.e. tiny problems for which the optimal solution can be computed quickly, we draw 1000 sample problems for 50 different values for $d$ and with process times and weights drawn uniformly at random from $\{1,..,100\}$ in each case.
Hence, each plot summarizes the results of 50000 experiments in which the optimal solution and the WSPT solution was computed.

The plots indicate that in the great majority of the cases, the WSPT solution deviates by at most 5\% from the optimal solution.
It is only a minority of less than 10\% of the cases in which gaps of up to 20\% occur.
However, these gaps seem to become smaller with an increasing number of jobs.
While in the 4 job case, gaps of even above 20\% can be observed, in the 8 and 9 job scenarios, even a gap of 10\% is hardly ever reached.
It is also noteworthy that the deviation seems to follow a hill-like structure with a focus in the first half, indicating that the problems in which the deadline is roughly between 10\% and 60\% of the total processing time exhibit the largest gaps.

Of course, this maybe surprising observation does not mean that it is not worthwhile to study dedicated algorithms.
First, there are indeed some cases in which the observed gap is substantially larger, and if, by chance, the problem at hand is of such a type, it would be odd to not have an appropriate algorithm available.
Second, the above discussion holds only for the single machine case.
In fact, a so intense analysis with computation of the optimal solution in each configuration is not feasible for the multi-machine case.
However, we will also see in Section \ref{sec:evaluation:finalresults} that the gap increases with an increasing number of machines (as opposed to an increasing number of jobs).
Therefore, we can certainly conclude that the WSPT ordering is a good advisor for search but that we can also improve upon it; sometimes significantly.

\section{Solution Approaches}
\label{sec:approaches}

\subsection{WSPT-Greedy and Naive Approaches}
\label{sec:approaches:naive}
The observations from Section \ref{sec:analysis:empirical} suggest a naive approach that simply ignores the deadline $d$, i.e. that pretends that the latest arrival time is $d = 0$.
For this case, the solution is simply WSPT for the one-machine case or equivalent to the weighted flowtime minimization for several machines without further constraints or conditions.
In the latter situation, we adopt the B\&B search presented in \cite{sarin1988improved}.
Since there is, by Lemma \ref{theorem:norepairpossible}, no way of efficiently repairing these solutions, we simply leave them as is.
This approach cannot be expected to be optimal most of the times, but it gives a reasonable baseline over which we certainly should be able to improve.

For the case of more than one machine, we can also think of other even simpler techniques to construct at least preliminary solutions.
Building on top of WSPT, we can use for example round robin (RR) or the ``earliest available machine'' (or first free, hence FF) rule.
Denoting the WSPT schedule as $\pi$, the RR procedure simply assigns job $\pi(j)$ to machine $\pi(j) \mod m$, and FF assigns $\pi(j)$ to the machine that gets available first (using lexicographical ordering as a tie-breaker).

\subsection{A Genetic Algorithm}
Thanks to the insight of Corollary \ref{theorem:yieldspartitionproblem}, the encoding of genomes for a genetic algorithm is straight forward.
For each job, we use an integer field with values in $\{1,..,m\}$ for the machine assigned to the job and a binary bit declaring whether it arrives strictly prior to the deadline or not, yielding genomes of length $2n$.
We will call the binary bit for the pre-deadline arrival of each job $j$ the \emph{priority} bit; it is set to 1 if the job arrives with priority (prior to $d$).

Note that most genomes are either invalid or dominated but can be repaired efficiently via random operations.
For some schedule, let $J_i$ be the jobs assigned to machine $i$ in the genome encoding, and let \jobsPriorityWithIndex{i} be the jobs arriving earlier and \jobsLateWithIndex{i} be the jobs arriving not earlier than deadline $d$ on that machine.
The genome is invalid if there is a machine $i$ such that at least one of the jobs in \jobsPriorityWithIndex{i} cannot start earlier than the deadline; formally $\sum_{j\in \jobsPriorityWithIndex{i} \setminus \{\hat{j}\}} p_j \geq d$ for every $\hat{j} \in \jobsPriorityWithIndex{i}$.
By ordering the jobs in \jobsPriorityWithIndex{i} by their processing time and randomly drawing a sequence of their indices, we can simply switch priority-bits to 0 in the order of the randomly drawn positions until the remaining jobs except the last one have a total processing time less than $d$.
This can be done efficiently yielding a total repair time of $\mathcal{O}(m \max_i |\jobsLateWithIndex{i}|)$.
Likewise, we can \emph{fill up} a set \jobsPriorityWithIndex{i} with jobs from \jobsLateWithIndex{i} until the same criterion is met with complexity $\mathcal{O}(m \max_i |J_i|)$.

While there are arbitrary mechanisms for cross-over and mutation imaginable, in this paper we adopt the following algorithm of creating offsprings.
W do not use cross-overs but only mutations of single individuals, which works as follows:
\begin{enumerate}
    \item \label{algo:genetic:samplemachines} Uniformly sample an integer $\xi \in \{1,..,MAX\}$ and then uniformly draw $\xi$ machines, where $1 \leq MAX \leq m$ is a parameter to limit the maximum changes per iteration.
    MAX is an upper bound for the number of machines on which changes may occur in a single mutation.
    A machine being selected means that it participates in the mutation; denote this set as $\hat{I}$.
    
    \item \label{algo:genetic:sampleunprioritizedjobs} For each machine $i \in \hat{I}$, uniformly sample \emph{one} prioritized job $j \in \jobsPriorityWithIndex{i}$ whose priority bit will be swapped to 0.
    Let $\hat{J}= \cup_{i\in \hat{I}}\jobsLateWithIndex{i}$ be the set of jobs on the participating machines that are currently not prioritized.
    
    \item After swapping one job for each $i\in \hat{I}$, the schedule for each of those machines is dominated in the sense discussed above, and we now conduct a \emph{fill-up} for each machine using the jobs in $\hat{J}$ until each machine is non-dominated but still valid.
    This fill-up is conducted sequentially for each machine $i$ in separation:
    \begin{enumerate}
        \item Determine the set of jobs $\hat{J}' \subseteq \hat{J}$ that can be located on machine $i$ with priorization such that all currently prioritized jobs on $i$ (except the swapped one) can still start prior to the deadline.
        \item If $\hat{J}' = \emptyset$, finish the mutation process for this machine.
        \item \label{algo:genetic:forbidmachinechanges} With a probability of 90\% remove all jobs that are not currently allocated on $i$ from this set unless this would yield an emptyset.
        \item Uniformly draw one of these jobs, set its priority bit to 1, and remove it from $\hat{J}$.
        \item repeat this sub-routine
    \end{enumerate}
\end{enumerate}
Some remarks are due on this algorithm.
\begin{itemize}
    \item First, this algorithm mutates, by construction, each valid genome to a non-dominated and valid new genome.
Hence, if we make sure that we have an initial population with only valid (not necessarily non-dominated) individuals, we do not need any additional repair mechanisms but always have only non-dominated and valid individuals in the population.
    \item Second, the fact that $\hat{J}$ is defined \emph{across} machines allows to swap jobs from one machine to another.
    
    \item Third, the algorithm fixes some constants in a rather conservative way to avoid overly intense mutations.
    On one hand, by (\ref{algo:genetic:samplemachines}+\ref{algo:genetic:sampleunprioritizedjobs}) it allows for \emph{at most} one prioritized job of each machine to be unprioritized in each mutation.
    On the other hand, step (\ref{algo:genetic:forbidmachinechanges}) limits the swapping of jobs across machines in a rather strict way.
    The motivation for these rather rigid mutations is due to the recognition that optimal solutions are probably not \emph{too} far away from WSPT solutions and hence one should not move too much in the space.
    A lot of changes in a single mutation increase the risk that helpful changes are undone or skipped due to other changes within the same mutation.
    Hence, we prefer to make small steps in the search space.
\end{itemize}
To escape from local minima, we consider a restart-technique that restarts the algorithm if no improvement has occurred in a pre-defined number of iterations.
Since we only work with integers, the improvements cannot converge to 0.

\subsection{Iterative Local Search}
Alternatively to the genetic algorithm, which, apart from the restarts, hardly undertakes serious attempts of leaving from local minima, we consider an iterative local search (ILS).
We use the same encoding scheme for the genomes as for the genetic algorithm but adopt a different strategy in traversing the space of those genomes.

The corresponding algorithm is depicted in Alg.\ref{algo:ils} (ILS).
First, we initialize a single solution, perhaps using the naive approaches above.
Then, we iteratively conduct random walks from that solution of a random length and try to locally improve from the schedule where the random walk led us for a pre-defined number of steps.
This cycle is conducted until no more time is left.

\begin{algorithm}[t]
$s \gets$ initialize solution\;
$s^* \gets s$\;
\While{there is time left}{
    \BlankLine
    \tcc{conduct a random walk}
    $i \gets 0$\;
    $s' \gets s$\;
    \For{$i < rand(k)$}{
        $s' \gets $\sc{getNeighbor}($s'$)\;
    }
    \BlankLine
    \tcc{locally improve}
    $i \gets 0$\;
    \For{$i < l$}{
        $s'' \gets $\sc{getNeighbor}($s'$)\;
        \If{$\phi(s'') \leq \phi(s')$}{
        $s' \gets s''$\;
        }
    }
    \If{$\phi(s') < \phi(s^*)$}{
    $s^* \gets s'$\;
    }
}
\caption{ILS}
\label{algo:ils}
\end{algorithm}

Note that {\sc getNeighbor} is a function that derives a new genome from a given one and as such can be seen as a mutation, and all mutations applicable for the genetic algorithm can be used here as well.
For comparability, we will use the same mutation as used in the genetic algorithm.

In a sense, ILS can be seen as a special case of the genetic algorithm above with pool size 1.
Every iteration of the main cycle can be seen as a restart of the GA, and instead of using an entirely random initialization, we use a random walk to generate the initial solution.
Since we only consider a single candidate at each point of time, this corresponds to a pool size of 1 and can, hence, be seen as a focused variant of the above GA.
An alternative approach we leave for future work would be to define a rather narrow (deterministic) environment for each solution and to consider \emph{all} neighbors in this environment and choose the best one.

\subsection{MILP Approach}
Our problem can be phrased as a mathematical optimization problem.
While it is well-known that MILP (Mixed Integer Linear Programming) based approaches typically cannot compete with other solutions due to complexity issues, MILP formulations are a standard notation and we hence nevertheless present the MILP for the sake of completeness.
Our formulation is in the manner of the linear ordering type \cite{Unlu2010}:\\[.5em]
\noindent
\emph{Sets}\\
$J$ set of jobs indexed in $j$ and $k$\\
$I$ set of parallel machines indexed in $i$\\

\noindent
\emph{Variables}\\
$C_j$: Completion time of job $j$. \\
$r_j$: variable release date of job $j$.\\
$z_{ij}$: 1 if job $j$ is processed on machine $i$; 0 otherwise.\\
$y_{jk}$: 1 if job $j$ is processed before job $k$; 0 otherwise.\\
$\hat{r}_k$: 1 if job $j$ is released before the deadline $d$; 0 otherwise.\\

\noindent
The parameters $w_j$, $p_j$, and $d$ for the weights, processing times, and the global deadline respectively were already defined in Sec. \ref{sec:problem}.
The goal is then to find

{
\footnotesize
$$\min_{C_j, j \in J} \sum_{j\in J} w_j(C_j-r_j)$$
subject to
\setcounter{equation}{0}
\begin{align}
    \sum_{i \in I} z_{ij} = 1  & & & \forall j \in J\\
    C_j \geq p_j + r_j & & & \forall j \in J\\
    C_k - p_k \geq C_j - M(3 - y_{jk} -z_{ij} -z_{ik}) & & & \forall j<k \in J, i \in I\\
    C_j - p_j \geq C_k - M(2 + y_{jk} -z_{ij} -z_{ik}) & & & \forall j<k \in J, i \in I\\
    r_j \leq d & & & \forall j \in J\\
    y_{jk} \in \{0,1\} & & & \forall j,k \in J, j< k\\
    z_{ij} \in \{0,1\} & & & \forall k \in J, i \in I\\\midrule
    d - r_j \leq M (1 - \hat{r}_j) & & & \forall j \in J\\
    r_j - d \leq M \hat{r}_j & & & \forall j \in J\\
    \frac{p_j}{w_j} \leq \frac{p_k}{w_k} + M(4-y_{jk} - z_{ij} -z_{ik} - \hat{r}_j) & & &\forall j<k \in J, i \in I\\
    \frac{p_k}{w_k} \leq \frac{p_j}{w_j} + M(3 + y_{jk} - z_{ij} -z_{ik} - \hat{r}_k) & & & \forall j<k \in J, i \in I\\
    \hat{r}_{j} \in \{0,1\} & & & \forall j \in J
\end{align}
}
The restrictions above the line are standard for this kind of problem.
Constraint set (1) establishes that a job must be processed in exactly one machine. Constraint set (2) establishes that the completion time of a job must be at least its release date plus its processing time. Constraint sets (3) and (4) ensure that two jobs cannot be simultaneously processed on the same machine. Constraint set (5) ensures that no job can be release after the deadline.
Constraint sets (6) and (7) correspond to the domain of the variables.

The restrictions below the line can be used to incorporate the knowledge that jobs arriving not earlier than $d$ must follow the WSPT rule as per Lemma \ref{theorem:swpt}.
Constraint sets (8) and (9) relate the release date variables $r_j$ with the binary variable $\hat{r}_k$ used to indicate that a job arrives exactly at time $d$.
Constraint sets (10) and (11) ensure that the jobs on each machine arriving not earlier than $d$.
The last constraint set (12) makes $\hat{r}_j$ a binary variable.

It is not immediately clear whether the net benefit of considering the knowledge of WSPT ordering at the cost of these additional variables is positive.
In the end, they make the model more complicated.
In Sec. \ref{sec:evaluation:finalresults}, we compare the performance of CPLEX with and without this constraint set.

\subsection{CP Approach}
A more elegant encoding is possible by using constraint programming (CP).
A typical property of constraint programming languages is the support of \emph{interval types}, which make them particularly amenable to scheduling problems, in which the time of a job spent on a machine corresponds to such an interval.
In particular, the OPL language \cite{van1999opl} supports build-in constraints for alternatives and no-overlap on such interval types.

In contrast to the MILP formulation, we have far fewer decision variables.
In fact, we only need one interval-valued decision variable $z_{ij}$ for each combination of jobs and machines.
Other than in the MILP formulation, $z_{ij}$ is an \emph{optional} variable (can be set to an interval, which will correspond exactly to the time of job $j$ on machine $i$, or to \emph{null} if job $j$ is not processed on machine $i$).
Intuitively, we will require that for every job $j$ exactly one machine $i$ is chosen (all other $z_{i'j}$ are null) and that the intervals on any machine $i$ are chosen so that they do not overlap.

In addition to this decision variable, we can derive several expressions that help us formulate the problem.
Let $s_j, e_j$, and $r_j$ be the starting time, end time, and release date of job $j$ respectively.
Note that these variables can be easily \emph{derived} from the decision variables $z_{ij}$ and do not constitute, in contrast to the MILP model, decision variables themselves.
Then we simply add the constraints

{
\footnotesize
\setcounter{equation}{0}
\begin{align}
    alternative&(z_{1j},..,z_{mj}) & \forall j \in J\\
    noOverlap&(z_{i1}, ..,z_{in}) & \forall i \in I\\\midrule
    z_{ij}\land z_{ik} \land (s_j < s_k) \land (s_j \geq d) &\rightarrow \frac{p_j}{w_j} \leq \frac{p_k}{w_k} & \forall i \in I
\end{align}
}
The impact of the constraints is straight forward.
Condition (1) assures that each job is placed on exactly one machine.
Condition (2) assures that the jobs on each machine do not overlap, and condition (3) requires that the jobs on any machine $i\in [m]$ arriving not earlier than the deadline are ordered by WPST (Lemma \ref{theorem:swpt}).
Note that while the first two constraints are necessary to produce \emph{valid} solutions, this last constraint serves as an \emph{additional information} for the solver to early prune necessarily sub-optimal solutions.
In order to assess the advantage of this modeling technique, again, we evaluate the performance of the CPSolver with and without constraint (3).

\subsection{Monte Carlo Tree Search}
Monte Carlo Tree Search (MCTS) \cite{browne2012survey} is an algorithm framework that is typically applied in adversarial or non-deterministic setups such as games.
The goal is to semi-randomly draw path in a given tree with the goal in order to derive a so-called \emph{policy} that is able to reach good tree regions on average.
The classical realm of applying MCTS are scenarios that can be described by Markov Decision Processes (MDPs), but its application to tree search scenarios in which traditional heuristic approaches do not work due to the limited ability of intermediate node evaluations has also been considered in the context of single player games \cite{schadd2008single}.
A similar technique has been applied in the context of Automated Machine Learning \cite{mohr2018ml}.

The application of MCTS to this scheduling problem is interesting, because classical Branch and Bound (B\&B) or heuristic techniques are, at this point, not applicable.
B\&B and heuristic search are the two principal players in tree-based search.
As discussed in detail in Sec. \ref{sec:analysis:theoreticalinsights}, the main obstacle for the use of B\&B is that no meaningful lower bounds are known, and that it is entirely unclear how to transfer the bounds for the case without a latest arrival time to this problem.
The NP-hardness of even the one-machine case indicates that this might not be possible in general.
A very similar argument applies for the case of heuristic search: To make such a search effective, we would need a good lower bound for the partial evaluation of each partial schedule.
In the absence of such a lower bound, no efficient heuristic search is possible.

In this paper, we consider a version of MCTS that is based on Thompson sampling \cite{bai2013bayesian}.
This approach pursues Bayesian inference, assuming a normal distribution of observations and applying Normal-Gamma as the corresponding prior distribution.
The approach relies on an incremental fit of a Dirichlet distribution to learn the underlying transition probabilities, which is however not necessary in our deterministic context.
Based on the three components (Dirichlte + Normal(+ Gamma)), the approach is called DNG.
In several preliminary tests we found DNG to be strictly superior to the much more famous and theoretically more underpinned algorithm UCT (short for Upper Confidence Bounds) \cite{kocsis2006bandit}, which is why we use it here.
Describing the formal details of the algorithm is beyond the scope of this paper; we refer the interested reader to \cite{browne2012survey,bai2013bayesian} for more details.

The tree searched by this algorithm is defined by iteratively deciding for each operation the machine it will be allocated at and whether it will be allocated with priority.
Every inner node is associated with the next operation to be assigned and has one successor for each machine for the case that it should be prioritized (if possible) or not.
In this sense, we use a tree based approach to enumerate the different genome encodings already used for the local searches.
At this stage, we do not adopt a specific technique to break symmetries.

\section{Experimental Evaluation}
In this evaluation, we will compare the above algorithms with each other.
To our knowledge, this is the first paper to address the described problem, so there are no other approaches to compare our solutions against.
Also, we are currently not aware of meaningful lower bounds (and due to the difficulties discussed in Sec. \ref{sec:analysis:theoreticalinsights} there are no obvious candidates in sight), and the bounds produced by relaxation of the MILP are off the mark.
Hence, we will express the solution qualities in terms of their gap to the best in-portfolio solution.

In the following, we proceed in three steps.
First, we describe the exact experiment setup.
Second, we discuss the results obtained after the instance-specific timeouts.
Third, we analyze the behavior of the algorithms in terms of advances in the best known solution up to a given point of time.
The complete code (including the original result tables) for these experiments are available for public\footnote{\url{https://github.com/fmohr/jobscheduling}}.

\subsection{Experiment Setup}
\label{sec:evaluation:setup}
We consider the following setup of synthetically generated problems.
In accordance with \cite{Nessah2008}, we generate experiments for each number of machines $m \in \{1, 2, 3, 5, 9\}$, and
    jobs $n = |J| \in \{$20, 30, 40, 60, 80, 100, 120, 140, 160, 180, 200, 300, 400, 500$\}$, i.e.
$5\times 14$ = 70 instance sizes in total.
Processing times and weights are drawn uniformly from $\{1,..,100\}$ in all instances.
For each such setup, 8 deadlines are considered, one for each 10\% interval border of the total processing time of the respective setup divided by the number of machines.
To reduce the influence of random effects, for each instance size, we consider 50 uniformly drawn samples.
This yields a total number of $70\cdots 8 \cdots 50 = 28000$ problem instances.

The timeout of the experiments depends on both the number of jobs and machines.
More precisely, we grant a total runtime of $4mn$ seconds to each experiment, where $m$ is the number of machines and $n$ is the number of jobs.
Hence, the smallest experiment has a timeout of $4s\cdot 1 \cdot 20 = 80$s, whereas the largest one has a timeout of $4s\cdot9\cdot500=18000s=5$h.
These timeouts are admittedly arbitrary, but, to compensate for this, we will add a discussion of the behavior of the different algorithms \emph{during} runtime.
For the execution of a single algorithm on each of the 28000 problem instances, this yields a total timeout of $9320000s \approx 108$ days.

Each problem instance is solved with the eleven proposed algorithms:
The naive approach (naive, which is WSPT for $m$=1 and Branch \& Bound \cite{sarin1988improved} for $m > 1$), MCTS (mcts), GA with random initialization (ga-random), GA with one genome being the round robin initialization (cf. \ref{sec:approaches:naive}) and the rest random (ga-rr), GA with one genome being the first-free initialization (cf. \ref{sec:approaches:naive}) and the rest random (ga-ff), GA with one genome for RR and FF respectively and the rest random (ga-rr-ff), ILS (ils), Ilog CPlex 12.10 on the MILP encoding with/without WSPT constraints (cplex+/cplex-), Ilog CP Solver 12.10 on the constraint programming problem with/without the WSPT constraints (cp+/cp-).
This yields a total of 308 thousand experiments with a total timeout of $3.25$ years.

To carry out these experiments in acceptable time, the runs were conducted in a compute center with the following technical specification of the individual processes.
Every process had access to one main CPU for essential computations and three buffer CPUs for thread maintenance and garbage collections.
In other words, no essential parallelization techniques were adopted in order to not give such implementation-specific advantages to any of the algorithms and to increase comparability among them.
The CPU architecture was Intel Xeon E5-2670, 2.6Ghz.
Each process was granted 16GB of which effectively only 8GB were assigned to the Java Virtual Machine (JVM) and the rest to the java process itself (to handle meta-operations in the process).
Within this specification, up to 450 jobs could be executed in parallel, yielding a total runtime of approximately one week.

All algorithms except CPLEX and CPSolver have been implemented in Java based on the Java library for artificial intelligence \emph{AILibs}\footnote{\url{https://github.com/starlibs/AILibs}}.
In order to maximize reproducibility, the code together with the experiment configuration is available at the code repository coming along with this paper indicated above.

\subsection{Analysis of Final Results}
\label{sec:evaluation:finalresults}
In this section, we will answer the following three research questions:
\begin{enumerate}
    \item[RQ 1] Is there a single best algorithm that comprehensively outperforms the others?
    \item[RQ 2] What improvement is possible over the Naive approach and how does this depend on the deadline?
    \item[RQ 3] What is the effect of the deadline on the variance in the results?
    \item[RQ 4] Do CPLEX or the CPSolver benefit from adding the WSPT constraints to the problem formulation?
\end{enumerate}
We answer each question in a separate subsection.

\subsubsection{Best Solution}
\begin{table*}[t!]
\centering
    \resizebox{\textwidth}{!}{%
    \bgroup
\setlength\tabcolsep{3pt}
\renewcommand\arraystretch{.0}
\begin{tabular}{l|rr|rr|rr|rr|rr|rr|rr|rr|rr|rr|rr}
\toprule
    \multicolumn{1}{c|}{Size} &  \multicolumn{2}{|c|}{Naive} &  \multicolumn{2}{|c|}{MCTS} &  \multicolumn{2}{|c|}{ga-random} &  \multicolumn{2}{|c|}{ga-rr}&  \multicolumn{2}{|c|}{ga-ff} &  \multicolumn{2}{|c|}{ga-rr-ff} &   \multicolumn{2}{|c|}{ils} & \multicolumn{2}{|c|}{cplex+} & \multicolumn{2}{|c|}{cplex-} &     \multicolumn{2}{|c|}{cp+} & \multicolumn{2}{|c}{cp-} \\
\midrule
  1 x 20 &              0.5\% &   (1.9\%) &   \underline{0.4\%} &   (1.9\%) &   \underline{0.1\%} &      (4.2\%) &     \textbf{0.0\%} &  (2.0\%) &     \textbf{0.0\%} &  (1.2\%) &     \textbf{0.0\%} &           (2.0\%) &     \textbf{0.0\%} &           (0.2\%) &               1.1\% &          (14.9\%) &               1.2\% &          (23.3\%) &     \textbf{0.0\%} &      (1.2\%) &  \textbf{0.0\%} &  \textbf{(0.0\%)} \\
  1 x 30 &              0.5\% &   (1.4\%) &               0.6\% &   (4.7\%) &               0.7\% &     (11.5\%) &  \underline{0.1\%} &  (1.9\%) &              0.1\% &  (1.9\%) &  \underline{0.1\%} &           (1.9\%) &     \textbf{0.0\%} &           (0.3\%) &   \underline{8.9\%} &          (34.9\%) &              12.9\% &          (27.3\%) &     \textbf{0.0\%} &  ($\infty$) &  \textbf{0.0\%} &  \textbf{(0.0\%)} \\
  1 x 40 &              0.4\% &   (0.8\%) &               1.8\% &   (9.1\%) &               1.9\% &     (21.6\%) &              0.1\% &  (1.0\%) &              0.2\% &  (1.0\%) &  \underline{0.2\%} &           (1.0\%) &     \textbf{0.0\%} &           (0.1\%) &              19.2\% &         (108.2\%) &              13.7\% &          (25.6\%) &     \textbf{0.0\%} &  ($\infty$) &  \textbf{0.0\%} &  \textbf{(0.0\%)} \\
  1 x 60 &              0.1\% &   (0.3\%) &               3.2\% &  (10.0\%) &               4.7\% &     (29.7\%) &              0.1\% &  (0.5\%) &              0.1\% &  (0.5\%) &  \underline{0.1\%} &           (0.5\%) &     \textbf{0.0\%} &           (0.1\%) &              23.4\% &          (82.9\%) &              22.5\% &         (147.5\%) &     \textbf{0.0\%} &  ($\infty$) &  \textbf{0.0\%} &  \textbf{(0.0\%)} \\
  1 x 80 &  \underline{0.1\%} &   (0.2\%) &   \underline{2.9\%} &  (12.7\%) &  \underline{12.9\%} &     (36.5\%) &  \underline{0.1\%} &  (0.2\%) &  \underline{0.1\%} &  (0.2\%) &  \underline{0.1\%} &           (0.1\%) &     \textbf{0.0\%} &           (0.0\%) &  \underline{40.6\%} &          (84.1\%) &  \underline{42.4\%} &          (54.1\%) &     \textbf{0.0\%} &  ($\infty$) &  \textbf{0.0\%} &  \textbf{(0.0\%)} \\
 1 x 100 &              0.1\% &   (0.1\%) &   \underline{3.1\%} &  (13.3\%) &   \underline{8.3\%} &     (29.4\%) &  \underline{0.1\%} &  (0.1\%) &     \textbf{0.0\%} &  (0.1\%) &     \textbf{0.1\%} &           (0.1\%) &     \textbf{0.0\%} &  \textbf{(0.0\%)} &  \underline{71.7\%} &         (136.5\%) &              67.9\% &         (127.9\%) &          $\infty$ &  ($\infty$) &  \textbf{0.0\%} &           (0.1\%) \\
 1 x 120 &              0.1\% &   (0.1\%) &   \underline{8.3\%} &  (16.3\%) &   \underline{0.4\%} &     (67.1\%) &     \textbf{0.0\%} &  (0.1\%) &     \textbf{0.0\%} &  (0.1\%) &     \textbf{0.0\%} &           (0.1\%) &     \textbf{0.0\%} &  \textbf{(0.0\%)} &              89.1\% &         (182.7\%) &              99.9\% &         (195.8\%) &          $\infty$ &  ($\infty$) &           0.1\% &           (0.7\%) \\
 1 x 140 &     \textbf{0.0\%} &   (0.1\%) &   \underline{9.5\%} &  (15.0\%) &               0.4\% &      (1.6\%) &     \textbf{0.0\%} &  (0.1\%) &     \textbf{0.0\%} &  (0.1\%) &     \textbf{0.0\%} &           (0.1\%) &     \textbf{0.0\%} &  \textbf{(0.0\%)} &             107.1\% &         (215.1\%) &             127.2\% &         (198.4\%) &          $\infty$ &  ($\infty$) &           0.5\% &           (4.2\%) \\
 1 x 160 &     \textbf{0.0\%} &   (0.1\%) &  \underline{11.2\%} &  (21.3\%) &   \underline{0.6\%} &      (2.4\%) &     \textbf{0.0\%} &  (0.1\%) &     \textbf{0.0\%} &  (0.1\%) &     \textbf{0.0\%} &           (0.1\%) &     \textbf{0.0\%} &  \textbf{(0.0\%)} &             122.6\% &         (245.2\%) &             135.4\% &         (245.2\%) &          $\infty$ &  ($\infty$) &           1.4\% &           (6.5\%) \\
 1 x 180 &     \textbf{0.0\%} &   (0.1\%) &  \underline{10.9\%} &  (17.1\%) &   \underline{0.6\%} &      (1.8\%) &     \textbf{0.0\%} &  (0.1\%) &     \textbf{0.0\%} &  (0.1\%) &     \textbf{0.0\%} &           (0.1\%) &     \textbf{0.0\%} &  \textbf{(0.0\%)} &             134.7\% &         (303.3\%) &             125.7\% &         (298.2\%) &          $\infty$ &  ($\infty$) &           2.1\% &           (5.8\%) \\
 1 x 200 &     \textbf{0.0\%} &   (0.1\%) &  \underline{11.8\%} &  (19.1\%) &   \underline{0.8\%} &      (2.2\%) &     \textbf{0.0\%} &  (0.1\%) &     \textbf{0.0\%} &  (0.1\%) &     \textbf{0.0\%} &           (0.1\%) &     \textbf{0.0\%} &  \textbf{(0.0\%)} &           $\infty$ &       ($\infty$) &           $\infty$ &       ($\infty$) &          $\infty$ &  ($\infty$) &           4.5\% &           (8.5\%) \\
 1 x 300 &     \textbf{0.0\%} &   (0.0\%) &  \underline{14.9\%} &  (21.0\%) &   \underline{1.2\%} &      (2.8\%) &     \textbf{0.0\%} &  (0.0\%) &     \textbf{0.0\%} &  (0.0\%) &     \textbf{0.0\%} &           (0.0\%) &     \textbf{0.0\%} &  \textbf{(0.0\%)} &           $\infty$ &       ($\infty$) &           $\infty$ &       ($\infty$) &          $\infty$ &  ($\infty$) &          12.7\% &          (23.3\%) \\
 1 x 400 &     \textbf{0.0\%} &   (0.0\%) &  \underline{17.6\%} &  (25.2\%) &   \underline{1.6\%} &      (2.9\%) &     \textbf{0.0\%} &  (0.0\%) &     \textbf{0.0\%} &  (0.0\%) &     \textbf{0.0\%} &           (0.0\%) &     \textbf{0.0\%} &  \textbf{(0.0\%)} &           $\infty$ &       ($\infty$) &           $\infty$ &       ($\infty$) &          $\infty$ &  ($\infty$) &          21.9\% &          (30.7\%) \\
 1 x 500 &     \textbf{0.0\%} &   (0.0\%) &  \underline{18.3\%} &  (25.9\%) &   \underline{2.0\%} &      (3.3\%) &     \textbf{0.0\%} &  (0.0\%) &     \textbf{0.0\%} &  (0.0\%) &     \textbf{0.0\%} &           (0.0\%) &     \textbf{0.0\%} &  \textbf{(0.0\%)} &           $\infty$ &       ($\infty$) &           $\infty$ &       ($\infty$) &          $\infty$ &  ($\infty$) &          27.5\% &          (31.6\%) \\\midrule
  2 x 20 &              1.0\% &   (3.7\%) &               5.3\% &  (11.7\%) &               0.7\% &      (7.0\%) &              0.5\% &  (3.1\%) &              0.4\% &  (2.5\%) &  \underline{0.6\%} &           (3.1\%) &     \textbf{0.0\%} &           (1.0\%) &               0.5\% &           (2.6\%) &               0.8\% &           (2.5\%) &     \textbf{0.0\%} &      (1.2\%) &  \textbf{0.0\%} &  \textbf{(0.1\%)} \\
  2 x 30 &              0.9\% &   (2.4\%) &               8.8\% &  (14.3\%) &               1.5\% &     (10.3\%) &              1.0\% &  (4.9\%) &              0.7\% &  (3.1\%) &  \underline{0.7\%} &           (3.1\%) &     \textbf{0.0\%} &           (0.9\%) &               4.4\% &          (26.9\%) &               5.6\% &          (12.5\%) &     \textbf{0.0\%} &      (2.0\%) &           0.1\% &  \textbf{(0.4\%)} \\
  2 x 40 &              0.8\% &   (1.4\%) &              11.4\% &  (18.8\%) &               2.8\% &     (19.2\%) &              1.0\% &  (3.5\%) &              0.7\% &  (1.8\%) &  \underline{0.7\%} &           (1.7\%) &     \textbf{0.0\%} &           (0.8\%) &   \underline{6.9\%} &          (27.9\%) &               8.0\% &          (18.4\%) &     \textbf{0.0\%} &  ($\infty$) &           0.1\% &  \textbf{(0.2\%)} \\
  2 x 60 &              0.5\% &   (0.7\%) &              16.0\% &  (29.2\%) &               6.6\% &     (41.4\%) &              0.7\% &  (2.9\%) &              0.5\% &  (0.8\%) &              0.4\% &           (0.8\%) &     \textbf{0.0\%} &           (0.6\%) &              16.0\% &          (40.4\%) &              16.4\% &          (32.5\%) &          $\infty$ &  ($\infty$) &           0.1\% &  \textbf{(0.4\%)} \\
  2 x 80 &              0.4\% &   (0.5\%) &              10.1\% &  (27.2\%) &               6.5\% &     (15.7\%) &              0.7\% &  (1.7\%) &              0.4\% &  (0.5\%) &  \underline{0.3\%} &           (0.4\%) &     \textbf{0.0\%} &  \textbf{(0.1\%)} &              31.2\% &          (56.5\%) &              28.5\% &          (58.9\%) &          $\infty$ &  ($\infty$) &           0.3\% &           (0.4\%) \\
 2 x 100 &              0.3\% &   (0.4\%) &              11.1\% &  (31.6\%) &               8.6\% &     (28.1\%) &              0.3\% &  (2.3\%) &              0.2\% &  (0.4\%) &              0.2\% &           (0.4\%) &     \textbf{0.0\%} &  \textbf{(0.0\%)} &              36.4\% &          (53.3\%) &              41.4\% &          (85.4\%) &          $\infty$ &  ($\infty$) &           0.4\% &          (25.2\%) \\
 2 x 120 &              0.3\% &   (0.7\%) &              28.4\% &  (38.9\%) &               1.6\% &     (34.2\%) &              0.3\% &  (1.0\%) &              0.2\% &  (0.4\%) &              0.2\% &           (0.3\%) &     \textbf{0.0\%} &  \textbf{(0.0\%)} &              44.1\% &         (101.3\%) &              42.5\% &          (96.3\%) &          $\infty$ &  ($\infty$) &           0.7\% &           (4.0\%) \\
 2 x 140 &              0.2\% &   (0.3\%) &              30.7\% &  (38.2\%) &               1.6\% &      (4.2\%) &              0.3\% &  (1.1\%) &              0.1\% &  (0.2\%) &              0.1\% &           (0.3\%) &     \textbf{0.0\%} &  \textbf{(0.0\%)} &              68.7\% &         (235.7\%) &              71.8\% &         (115.6\%) &          $\infty$ &  ($\infty$) &           2.0\% &           (9.9\%) \\
 2 x 160 &              0.2\% &   (0.5\%) &              33.6\% &  (43.4\%) &               1.7\% &      (3.7\%) &              0.3\% &  (0.8\%) &              0.1\% &  (0.3\%) &              0.1\% &           (0.3\%) &     \textbf{0.0\%} &  \textbf{(0.0\%)} &              85.7\% &       ($\infty$) &              75.4\% &         (227.3\%) &          $\infty$ &  ($\infty$) &           5.9\% &          (11.8\%) \\
 2 x 180 &              0.2\% &   (0.2\%) &              34.9\% &  (50.9\%) &               1.7\% &      (4.0\%) &              0.2\% &  (1.1\%) &              0.1\% &  (0.2\%) &              0.1\% &           (0.2\%) &     \textbf{0.0\%} &  \textbf{(0.0\%)} &           $\infty$ &       ($\infty$) &           $\infty$ &       ($\infty$) &          $\infty$ &  ($\infty$) &           8.6\% &          (16.8\%) \\
 2 x 200 &              0.1\% &   (0.2\%) &              37.4\% &  (49.6\%) &               1.7\% &      (3.8\%) &              0.2\% &  (0.6\%) &              0.1\% &  (0.2\%) &              0.1\% &           (0.2\%) &     \textbf{0.0\%} &  \textbf{(0.0\%)} &           $\infty$ &       ($\infty$) &           $\infty$ &       ($\infty$) &          $\infty$ &  ($\infty$) &          11.2\% &          (19.4\%) \\
 2 x 300 &              0.1\% &   (0.1\%) &              43.4\% &  (53.8\%) &   \underline{2.0\%} &      (3.9\%) &              0.1\% &  (0.7\%) &     \textbf{0.0\%} &  (0.1\%) &     \textbf{0.0\%} &           (0.1\%) &     \textbf{0.0\%} &  \textbf{(0.0\%)} &           $\infty$ &       ($\infty$) &           $\infty$ &       ($\infty$) &          $\infty$ &  ($\infty$) &          21.9\% &          (27.4\%) \\
 2 x 400 &              0.1\% &   (0.1\%) &              45.5\% &  (56.5\%) &   \underline{2.2\%} &      (4.4\%) &  \underline{0.1\%} &  (0.5\%) &     \textbf{0.0\%} &  (0.1\%) &     \textbf{0.0\%} &           (0.1\%) &     \textbf{0.0\%} &  \textbf{(0.0\%)} &           $\infty$ &       ($\infty$) &           $\infty$ &       ($\infty$) &          $\infty$ &  ($\infty$) &          24.8\% &          (29.6\%) \\
 2 x 500 &     \textbf{0.1\%} &   (0.1\%) &              48.5\% &  (55.5\%) &   \underline{2.5\%} &      (4.1\%) &  \underline{0.1\%} &  (0.3\%) &     \textbf{0.0\%} &  (0.0\%) &     \textbf{0.0\%} &           (0.0\%) &     \textbf{0.0\%} &  \textbf{(0.0\%)} &           $\infty$ &       ($\infty$) &           $\infty$ &       ($\infty$) &          $\infty$ &  ($\infty$) &          28.1\% &          (32.8\%) \\\midrule
  3 x 20 &              1.8\% &   (4.6\%) &               8.3\% &  (16.2\%) &               0.7\% &      (5.4\%) &              0.8\% &  (4.8\%) &              0.8\% &  (4.0\%) &              0.7\% &           (3.4\%) &     \textbf{0.0\%} &           (1.5\%) &               0.5\% &           (2.2\%) &   \underline{0.6\%} &           (2.8\%) &  \underline{0.2\%} &      (1.4\%) &           0.5\% &  \textbf{(1.3\%)} \\
  3 x 30 &              1.4\% &   (2.6\%) &               8.8\% &  (15.2\%) &               2.0\% &     (23.5\%) &              1.4\% &  (8.3\%) &              0.9\% &  (2.9\%) &  \underline{1.1\%} &           (2.7\%) &     \textbf{0.1\%} &           (1.4\%) &               2.4\% &           (7.4\%) &               1.9\% &           (4.4\%) &              0.2\% &      (1.3\%) &           0.2\% &  \textbf{(1.3\%)} \\
  3 x 40 &              1.2\% &   (1.8\%) &              12.1\% &  (24.0\%) &               3.5\% &     (15.3\%) &              1.6\% &  (7.0\%) &              1.0\% &  (2.3\%) &              1.0\% &           (2.5\%) &     \textbf{0.0\%} &           (1.3\%) &               7.0\% &          (15.5\%) &               5.5\% &           (8.0\%) &              0.2\% &      (1.2\%) &           0.4\% &  \textbf{(1.2\%)} \\
  3 x 60 &              0.9\% &   (1.1\%) &              15.9\% &  (25.9\%) &               4.2\% &     (17.5\%) &              1.4\% &  (4.6\%) &              0.6\% &  (1.2\%) &              0.5\% &           (1.3\%) &     \textbf{0.0\%} &  \textbf{(0.9\%)} &              14.5\% &          (37.5\%) &              14.6\% &          (25.7\%) &              0.3\% &  ($\infty$) &           0.4\% &           (1.3\%) \\
  3 x 80 &              0.8\% &   (1.2\%) &               8.7\% &  (22.6\%) &               8.6\% &     (21.6\%) &              1.4\% &  (3.6\%) &              0.5\% &  (0.8\%) &              0.4\% &           (0.7\%) &     \textbf{0.0\%} &  \textbf{(0.3\%)} &              24.0\% &          (36.9\%) &              22.1\% &          (35.4\%) &          $\infty$ &  ($\infty$) &           0.4\% &           (0.8\%) \\
 3 x 100 &              0.5\% &   (0.8\%) &              11.6\% &  (36.7\%) &               7.6\% &     (19.1\%) &              1.3\% &  (4.3\%) &              0.3\% &  (0.5\%) &              0.3\% &           (0.5\%) &     \textbf{0.0\%} &  \textbf{(0.0\%)} &              26.4\% &          (52.6\%) &              27.0\% &          (45.3\%) &          $\infty$ &  ($\infty$) &           0.6\% &           (0.8\%) \\
 3 x 120 &              0.6\% &   (0.7\%) &              26.9\% &  (43.5\%) &               1.6\% &      (3.6\%) &              0.6\% &  (1.8\%) &              0.2\% &  (0.5\%) &              0.2\% &           (0.5\%) &     \textbf{0.0\%} &  \textbf{(0.0\%)} &              45.9\% &         (516.0\%) &              38.1\% &          (76.4\%) &          $\infty$ &  ($\infty$) &           0.5\% &           (0.8\%) \\
 3 x 140 &              0.5\% &   (0.6\%) &              28.9\% &  (40.8\%) &               1.6\% &      (3.7\%) &              0.5\% &  (2.7\%) &              0.2\% &  (0.3\%) &              0.2\% &           (0.4\%) &     \textbf{0.0\%} &  \textbf{(0.0\%)} &              79.3\% &       ($\infty$) &           $\infty$ &       ($\infty$) &          $\infty$ &  ($\infty$) &           0.6\% &           (1.2\%) \\
 3 x 160 &              0.5\% &   (0.8\%) &              31.3\% &  (40.5\%) &               1.6\% &      (3.3\%) &              0.5\% &  (2.1\%) &              0.2\% &  (0.3\%) &              0.2\% &           (0.3\%) &     \textbf{0.0\%} &  \textbf{(0.0\%)} &              88.4\% &       ($\infty$) &              92.1\% &         (185.3\%) &          $\infty$ &  ($\infty$) &           1.9\% &           (4.8\%) \\
 3 x 180 &              0.3\% &   (0.4\%) &              34.2\% &  (44.1\%) &               1.6\% &      (3.3\%) &              0.4\% &  (1.4\%) &              0.1\% &  (0.3\%) &              0.1\% &           (0.3\%) &     \textbf{0.0\%} &  \textbf{(0.0\%)} &           $\infty$ &       ($\infty$) &           $\infty$ &       ($\infty$) &          $\infty$ &  ($\infty$) &           3.1\% &          (29.0\%) \\
 3 x 200 &              0.4\% &   (0.4\%) &              35.0\% &  (45.5\%) &               1.7\% &      (3.5\%) &              0.5\% &  (1.4\%) &              0.1\% &  (0.2\%) &              0.1\% &           (0.3\%) &     \textbf{0.0\%} &  \textbf{(0.0\%)} &           $\infty$ &       ($\infty$) &           $\infty$ &       ($\infty$) &          $\infty$ &  ($\infty$) &           4.2\% &           (7.8\%) \\
 3 x 300 &              0.5\% &   (2.3\%) &              41.5\% &  (55.3\%) &               1.9\% &      (3.1\%) &              0.4\% &  (0.9\%) &     \textbf{0.1\%} &  (0.1\%) &              0.1\% &           (0.1\%) &     \textbf{0.0\%} &  \textbf{(0.0\%)} &           $\infty$ &       ($\infty$) &           $\infty$ &       ($\infty$) &          $\infty$ &  ($\infty$) &          15.0\% &          (27.0\%) \\
 3 x 400 &              0.2\% &   (0.2\%) &              45.7\% &  (55.4\%) &   \underline{2.2\%} &      (3.8\%) &  \underline{0.3\%} &  (1.0\%) &     \textbf{0.0\%} &  (0.1\%) &     \textbf{0.0\%} &           (0.1\%) &     \textbf{0.0\%} &  \textbf{(0.0\%)} &           $\infty$ &       ($\infty$) &           $\infty$ &       ($\infty$) &          $\infty$ &  ($\infty$) &          23.2\% &          (28.8\%) \\
 3 x 500 &              0.8\% &   (3.4\%) &              47.2\% &  (54.3\%) &   \underline{2.5\%} &      (3.6\%) &  \underline{0.2\%} &  (0.7\%) &     \textbf{0.0\%} &  (0.1\%) &     \textbf{0.0\%} &           (0.1\%) &     \textbf{0.0\%} &  \textbf{(0.0\%)} &           $\infty$ &       ($\infty$) &           $\infty$ &       ($\infty$) &          $\infty$ &  ($\infty$) &          26.0\% &          (32.2\%) \\\midrule
  5 x 20 &  \underline{2.1\%} &   (3.2\%) &   \underline{5.7\%} &  (19.9\%) &      \textbf{0.1\%} &      (0.6\%) &     \textbf{0.1\%} &  (1.6\%) &  \underline{0.3\%} &  (1.1\%) &              0.6\% &           (3.6\%) &     \textbf{0.1\%} &           (1.2\%) &               0.2\% &  \textbf{(0.5\%)} &      \textbf{0.1\%} &           (0.5\%) &              0.7\% &      (1.7\%) &           0.8\% &           (1.5\%) \\
  5 x 30 &              2.7\% &   (4.2\%) &               5.8\% &  (20.0\%) &               1.1\% &      (4.9\%) &              1.4\% &  (4.2\%) &              1.0\% &  (3.1\%) &              1.3\% &           (2.9\%) &     \textbf{0.0\%} &  \textbf{(0.1\%)} &               1.3\% &           (2.2\%) &               1.3\% &           (4.3\%) &              0.9\% &      (1.6\%) &           0.8\% &           (2.2\%) \\
  5 x 40 &              2.9\% &   (3.9\%) &               7.2\% &  (21.1\%) &               3.5\% &      (8.2\%) &              1.7\% &  (5.9\%) &              1.2\% &  (2.0\%) &              1.6\% &           (3.1\%) &     \textbf{0.0\%} &  \textbf{(0.6\%)} &               3.2\% &           (4.7\%) &               2.4\% &           (4.4\%) &              1.2\% &      (2.0\%) &           1.0\% &           (2.3\%) \\
  5 x 60 &              1.7\% &   (2.3\%) &               8.7\% &  (22.6\%) &               8.0\% &     (31.7\%) &              2.1\% &  (4.5\%) &              0.8\% &  (1.3\%) &              0.9\% &           (1.3\%) &     \textbf{0.0\%} &  \textbf{(0.4\%)} &               8.4\% &          (14.0\%) &               9.0\% &          (13.7\%) &              0.7\% &      (2.1\%) &           1.2\% &           (2.5\%) \\
  5 x 80 &              1.7\% &   (1.9\%) &               9.0\% &  (26.8\%) &               4.7\% &     (10.8\%) &              1.6\% &  (3.8\%) &              0.7\% &  (1.1\%) &              0.8\% &           (1.6\%) &     \textbf{0.0\%} &  \textbf{(0.2\%)} &              18.1\% &          (33.3\%) &              16.9\% &          (30.0\%) &          $\infty$ &  ($\infty$) &           1.3\% &           (2.2\%) \\
 5 x 100 &              1.4\% &   (2.3\%) &              10.3\% &  (30.1\%) &              10.0\% &     (32.5\%) &              1.8\% &  (4.4\%) &              0.5\% &  (0.9\%) &              0.5\% &           (1.1\%) &     \textbf{0.0\%} &  \textbf{(0.4\%)} &              22.2\% &          (33.1\%) &              20.0\% &          (26.1\%) &          $\infty$ &  ($\infty$) &           1.1\% &           (1.9\%) \\
 5 x 120 &              1.6\% &   (3.6\%) &              24.9\% &  (38.5\%) &               2.4\% &     (13.1\%) &              1.2\% &  (3.3\%) &              0.5\% &  (0.9\%) &              0.4\% &           (0.9\%) &     \textbf{0.0\%} &  \textbf{(0.5\%)} &              25.2\% &          (63.4\%) &              27.0\% &          (38.4\%) &          $\infty$ &  ($\infty$) &           0.9\% &           (1.3\%) \\
 5 x 140 &              1.3\% &   (1.8\%) &              27.0\% &  (37.1\%) &               2.3\% &     (11.6\%) &              1.0\% &  (2.3\%) &              0.3\% &  (0.7\%) &              0.3\% &           (0.7\%) &     \textbf{0.0\%} &  \textbf{(0.4\%)} &              30.0\% &          (65.6\%) &              29.4\% &          (45.3\%) &          $\infty$ &  ($\infty$) &           1.5\% &           (2.4\%) \\
 5 x 160 &              1.0\% &   (1.3\%) &              29.4\% &  (36.8\%) &               2.4\% &     (21.1\%) &              0.8\% &  (4.9\%) &              0.3\% &  (0.6\%) &              0.3\% &           (0.5\%) &     \textbf{0.0\%} &  \textbf{(0.2\%)} &              47.3\% &         (137.0\%) &              59.0\% &          (79.5\%) &          $\infty$ &  ($\infty$) &           1.6\% &           (2.6\%) \\
 5 x 180 &              1.1\% &   (1.2\%) &              31.0\% &  (45.0\%) &               2.6\% &     (76.5\%) &              0.8\% &  (2.9\%) &              0.2\% &  (0.5\%) &              0.2\% &  \textbf{(0.5\%)} &     \textbf{0.0\%} &           (0.6\%) &              77.0\% &       ($\infty$) &           $\infty$ &       ($\infty$) &          $\infty$ &  ($\infty$) &           2.0\% &           (2.5\%) \\
 5 x 200 &              0.9\% &   (1.1\%) &              33.1\% &  (39.1\%) &               2.5\% &     (30.3\%) &              0.8\% &  (1.8\%) &              0.2\% &  (0.4\%) &              0.2\% &  \textbf{(0.4\%)} &     \textbf{0.0\%} &           (0.5\%) &           $\infty$ &       ($\infty$) &           $\infty$ &       ($\infty$) &          $\infty$ &  ($\infty$) &           2.7\% &           (4.9\%) \\
 5 x 300 &              0.6\% &   (0.9\%) &              39.5\% &  (49.2\%) &               2.5\% &     (38.9\%) &              0.6\% &  (1.4\%) &              0.1\% &  (0.2\%) &              0.1\% &           (0.2\%) &     \textbf{0.0\%} &  \textbf{(0.1\%)} &           $\infty$ &       ($\infty$) &           $\infty$ &       ($\infty$) &          $\infty$ &  ($\infty$) &           8.0\% &          (13.0\%) \\
 5 x 400 &              0.7\% &   (0.8\%) &              43.8\% &  (49.8\%) &               2.6\% &     (41.6\%) &              0.5\% &  (1.5\%) &              0.1\% &  (0.2\%) &              0.1\% &           (0.2\%) &     \textbf{0.0\%} &  \textbf{(0.0\%)} &           $\infty$ &       ($\infty$) &           $\infty$ &       ($\infty$) &          $\infty$ &  ($\infty$) &          13.1\% &          (16.4\%) \\
 5 x 500 &              1.1\% &   (4.3\%) &              45.6\% &  (53.5\%) &               2.7\% &     (25.0\%) &              0.5\% &  (1.0\%) &              0.1\% &  (0.1\%) &              0.1\% &           (0.1\%) &     \textbf{0.0\%} &  \textbf{(0.0\%)} &           $\infty$ &       ($\infty$) &           $\infty$ &       ($\infty$) &          $\infty$ &  ($\infty$) &          25.0\% &          (27.7\%) \\\midrule
  9 x 20 &              1.0\% &   (5.7\%) &              25.3\% &  (55.0\%) &               0.1\% &      (5.8\%) &              0.1\% &  (2.2\%) &  \underline{0.2\%} &  (1.2\%) &              0.1\% &           (1.0\%) &     \textbf{0.0\%} &           (0.9\%) &      \textbf{0.0\%} &           (0.0\%) &      \textbf{0.0\%} &  \textbf{(0.0\%)} &     \textbf{0.0\%} &      (0.6\%) &  \textbf{0.0\%} &           (0.0\%) \\
  9 x 30 &              3.9\% &  (12.4\%) &              27.3\% &  (65.1\%) &               0.6\% &     (17.4\%) &              0.6\% &  (3.6\%) &  \underline{0.6\%} &  (3.3\%) &              0.5\% &           (3.1\%) &     \textbf{0.1\%} &           (1.3\%) &      \textbf{0.2\%} &           (1.0\%) &      \textbf{0.2\%} &  \textbf{(0.8\%)} &              0.8\% &      (1.7\%) &           0.8\% &           (1.5\%) \\
  9 x 40 &              4.2\% &  (11.4\%) &              27.6\% &  (36.1\%) &               0.6\% &      (1.7\%) &              0.6\% &  (1.6\%) &              0.7\% &  (1.7\%) &              0.8\% &           (1.8\%) &     \textbf{0.0\%} &  \textbf{(0.0\%)} &               1.1\% &           (2.5\%) &               0.9\% &           (1.4\%) &              2.2\% &      (4.9\%) &           1.7\% &           (3.2\%) \\
  9 x 60 &              4.4\% &   (5.5\%) &              27.0\% &  (44.7\%) &               1.8\% &      (3.1\%) &              1.7\% &  (3.0\%) &              1.3\% &  (2.5\%) &              1.3\% &           (2.4\%) &     \textbf{0.0\%} &  \textbf{(0.0\%)} &               3.9\% &          (11.0\%) &               4.4\% &           (6.2\%) &              1.6\% &      (3.2\%) &           2.1\% &           (3.2\%) \\
  9 x 80 &              3.3\% &   (3.7\%) &              27.4\% &  (38.9\%) &               2.5\% &      (4.5\%) &              1.8\% &  (3.8\%) &              1.3\% &  (2.0\%) &              1.3\% &           (2.1\%) &     \textbf{0.0\%} &  \textbf{(0.0\%)} &               8.5\% &          (18.9\%) &               9.1\% &          (17.3\%) &              1.5\% &     (27.4\%) &           2.1\% &           (3.0\%) \\
 9 x 100 &              3.5\% &   (4.9\%) &              28.6\% &  (37.0\%) &               2.7\% &     (39.8\%) &              1.7\% &  (2.7\%) &              1.0\% &  (1.6\%) &              1.0\% &           (1.7\%) &     \textbf{0.0\%} &  \textbf{(0.0\%)} &              14.1\% &          (25.0\%) &              13.2\% &          (19.3\%) &              3.6\% &  ($\infty$) &           1.7\% &           (2.8\%) \\
 9 x 120 &              3.0\% &   (3.7\%) &              29.6\% &  (38.3\%) &               3.0\% &      (6.0\%) &              1.6\% &  (2.9\%) &              0.7\% &  (1.3\%) &              0.8\% &           (1.3\%) &     \textbf{0.0\%} &  \textbf{(0.0\%)} &              18.1\% &          (29.5\%) &  \underline{16.9\%} &          (23.0\%) &              5.9\% &  ($\infty$) &           1.5\% &           (2.6\%) \\
 9 x 140 &              4.2\% &   (7.3\%) &              31.2\% &  (37.6\%) &               3.0\% &      (4.8\%) &              1.5\% &  (2.2\%) &              0.6\% &  (1.4\%) &              0.6\% &           (1.3\%) &     \textbf{0.0\%} &  \textbf{(0.0\%)} &  \underline{21.6\%} &          (37.1\%) &              22.0\% &          (39.8\%) &          $\infty$ &  ($\infty$) &           1.4\% &           (1.6\%) \\
 9 x 160 &              2.1\% &   (2.5\%) &              32.1\% &  (39.5\%) &               3.0\% &      (4.4\%) &              1.4\% &  (2.3\%) &              0.5\% &  (1.0\%) &              0.5\% &           (1.0\%) &     \textbf{0.0\%} &  \textbf{(0.1\%)} &  \underline{27.6\%} &          (53.4\%) &              30.7\% &          (45.4\%) &          $\infty$ &  ($\infty$) &           1.6\% &           (2.5\%) \\
 9 x 180 &              2.1\% &   (3.2\%) &              32.9\% &  (40.2\%) &   \underline{2.9\%} &      (5.0\%) &  \underline{1.1\%} &  (2.4\%) &              0.2\% &  (0.8\%) &              0.2\% &           (0.8\%) &     \textbf{0.1\%} &  \textbf{(0.6\%)} &           $\infty$ &       ($\infty$) &           $\infty$ &       ($\infty$) &          $\infty$ &  ($\infty$) &           1.5\% &           (2.3\%) \\
 9 x 200 &              2.4\% &   (3.8\%) &              33.1\% &  (42.9\%) &               3.1\% &      (4.9\%) &              1.3\% &  (2.8\%) &              0.4\% &  (0.6\%) &              0.4\% &           (0.8\%) &     \textbf{0.0\%} &  \textbf{(0.0\%)} &           $\infty$ &       ($\infty$) &           $\infty$ &       ($\infty$) &          $\infty$ &  ($\infty$) &           1.7\% &           (2.1\%) \\
 9 x 300 &              2.6\% &   (3.8\%) &              37.5\% &  (48.0\%) &   \underline{2.9\%} &      (4.7\%) &  \underline{0.9\%} &  (2.5\%) &     \textbf{0.1\%} &  (0.4\%) &     \textbf{0.0\%} &           (0.4\%) &  \underline{0.1\%} &  \textbf{(0.4\%)} &           $\infty$ &       ($\infty$) &           $\infty$ &       ($\infty$) &          $\infty$ &  ($\infty$) &           3.0\% &           (4.2\%) \\
 9 x 400 &              2.5\% &   (3.3\%) &              41.9\% &  (48.6\%) &               3.2\% &      (4.3\%) &              1.0\% &  (1.6\%) &              0.2\% &  (0.3\%) &              0.2\% &           (0.3\%) &     \textbf{0.0\%} &  \textbf{(0.0\%)} &           $\infty$ &       ($\infty$) &           $\infty$ &       ($\infty$) &          $\infty$ &  ($\infty$) &           7.5\% &          (10.9\%) \\
 9 x 500 &              4.2\% &  (13.0\%) &              45.6\% &  (50.2\%) &               3.4\% &      (4.8\%) &              0.9\% &  (1.7\%) &              0.1\% &  (0.2\%) &              0.1\% &           (0.2\%) &     \textbf{0.0\%} &  \textbf{(0.1\%)} &           $\infty$ &       ($\infty$) &           $\infty$ &       ($\infty$) &          $\infty$ &  ($\infty$) &          12.4\% &          (16.0\%) \\
\bottomrule
\end{tabular}

\egroup
    }

    \caption{Average and maximum final gaps of each algorithm compared to the best algorithm in the set.}
    \label{tab:results}
\end{table*}
For the sake of conciseness, we only present the results for the deadline at 40\% of the total processing time.
The full results can be found in the code repository coming along with this paper.
We chose the deadline of 40\% since it is the one that showed the largest gaps between the naive solution and the optimum in the one machine case (cf. Sec. \ref{sec:analysis:empirical}).

The results are shown in Table \ref{tab:results} in the following form.
For each pair of problems and algorithms, the table shows the \emph{mean} (\emph{maximum}) gap between the respective algorithm and the best observed performance.
That is, for each problem size, we consider the 100 solutions obtained by each algorithm on the 100 different instances sampled according to Sec. \ref{sec:evaluation:setup}.
For each instance, we compute the gap between each algorithm's score and the best score observed for this particular instance, which yields 100 gaps for each problem size and each algorithm.
Among these 100 gaps, we report the \emph{trimmed} mean (10\% cut) and the \emph{maximum} gap.
Intuitively, the figures then answer the following question: What is the mean gap (maximum gap ever) observed between the algorithm and the instance-wise best algorithm?
Bold entries show the \emph{best mean/max}, and mean scores are underlined if they are not statistically worse according to a Wilcoxon signed rank test at a p-level of 5\% (this significance test does not apply to maximum values, so there are no significance tests for the maxima).

According to our observations in Sec. \ref{sec:analysis:theoreticalinsights}, we do not compare the solutions against a lower bound since no meaningful lower bound is known at this point of time.
Again, we emphasize that we cannot directly re-use lower bounds on the problem without a deadline as discussed in \cite{sarin1988improved}, because the order on the machines does not generally obey the WSPT rule (only after the deadline).
The derivative of a high quality lower bound is clearly beyond the scope of this paper and is left for future work.

Maybe surprisingly, there is a clear winner in this comparison, which is the ILS algorithm.
There is only a single case in which the algorithm does not achieve the absolute best mean gap, and even in this case ILS is not statistically worse than the best one.
On the contrary, in the large majority of problems, ILS is statistically better than most or even all other algorithms.
Also, the extremely low maximum gaps indicate a high robustness of the ILS approach: in the vast majority of the cases, it has the lowest maximum gap, and even in the few cases where this is not so, its maximum gap is still small compared with some other sub-optimal algorithms.
It never exceeds the best maximum gap by even only 1\% point.

\subsubsection{Improvement over the Naive Approach}
\begin{figure}[t]
    \centering
    \includegraphics[width=.5\textwidth]{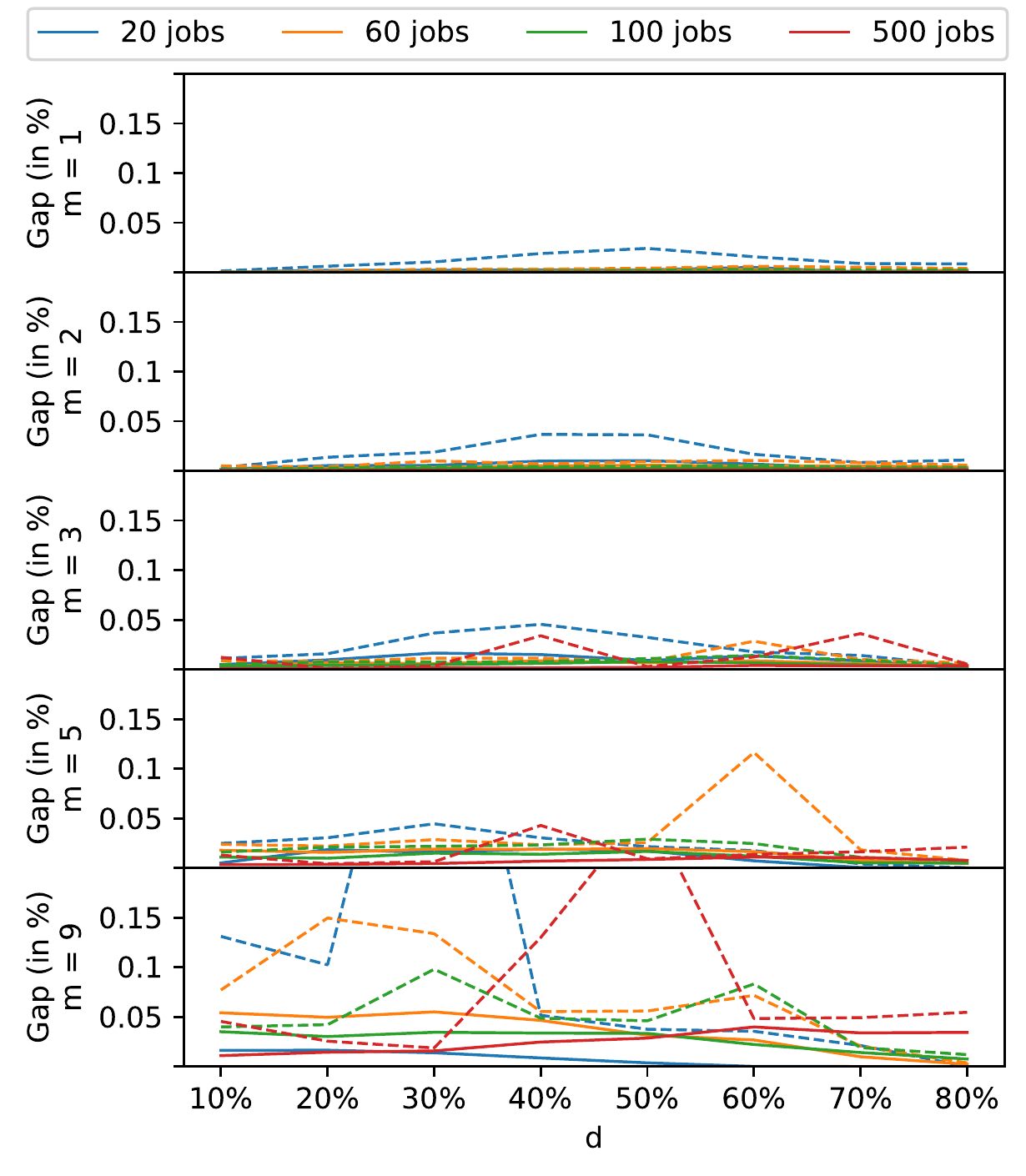}
    \caption{Improvement of ILS over the naive Approach. Solid lines are the mean improvement, dashed lines the maximum improvement.}
    \label{fig:results:improvementovernaive}
\end{figure}

Motivated by the experimental findings in Section \ref{sec:analysis:empirical}, the second question asks for the potential gains over a naive approach in scenarios with more than one machine (and with higher numbers of jobs).
The experimental results relevant to answer this question are depicted in Table \ref{tab:results} and Fig. \ref{fig:results:improvementovernaive}.
Since ILS is a dominating algorithm, we compare the naive approach here specifically against the best performer.
Aggregating over the 50 sample runs, the solid lines depict the \emph{mean} gaps observed for the different deadline indices, and the dashed lines depict the \emph{maximum} gap observed.
For readability, we only include the four problems for 20, 60, 100, and 500 jobs respectively.

There are two main observations: the first one is that the trend that the gap becomes smaller with an increasing number of jobs (cf. \ref{sec:analysis:empirical}) is also visible in the problems with more than one machine.
In fact, the top chart indeed suggests that this gap even \emph{vanishes} since even the observed maximum gap tends to 0.
However, this interpretation should be considered with care, because we are in fact \emph{not} comparing with the true optimum but against the best approach, and it is very likely that the gap of ILS to the true optimum also increases with an increasing number of jobs.
Therefore, this trend should not be overemphasized at this point.
The second is that, for a fixed number of jobs, the gap \emph{increases} with an increasing number of machines approximately for every deadline.

The fact that the solutions of the naive approach are very good has two possible explanations.
The first is that the solutions are in fact near-optimal, and that there is hence simply no potential to improve a lot upon them.
The second is that the suggested algorithms simply were not able to make stronger improvements.
While the observations in Sec \ref{sec:analysis:empirical} and the fact that a set of quite heterogeneous algorithms is not able to strongly improve upon the naive approach argue in favor of the first explanation, the observation that ILS is the only approach systematically superior to the others might argue in favor of the second explanation, since we can surely imagine to develop better versions of the ILS algorithm.

\subsubsection{Influence of the Deadline Parameter}
To answer the third research question, we can consider again the charts in Fig. \ref{fig:results:improvementovernaive}.
In line with the observations on the ground truth optimum solutions in Sec. \ref{sec:analysis:empirical}, we can observe that the gap between the naive solution and the ILS solution (serving as a surrogate for the optimum here) tends to be highest for intermediate values of deadlines in comparison to the overall deadline.

However, especially for the case with more machines, it seems that there is more interaction going on between the different parameters.
Specifically, here it seems that for a small number of jobs the problem is more difficult when the deadline is closer to 0.

At this point, the only valid conclusion is that the role of the deadline in general is not entirely clear and probably interacts with the other problem parameters such as number of jobs and machines.
This being said, we can still empirically observe that in rather small problems, deadlines close to 0 or the trivial upper bound tend to make the problem very similar to the problem without any deadline.
In such situations, if no implementation of an algorithm specialized on \probGraham is available, it seems not unreasonable to simply use an exact algorithm for \probGrahamNoDeadlines instead.

\subsubsection{The Effect of WSPT Rules on MILP and CP}
With respect to the CPLEX solver, we cannot determine a particular advantage of either formulation.
There are a dozen of cases in which cplex+ wins over cplex- and vice versa.
Among all the scenarios, there are only two case (the 3x120 and 5x180 scenarios) in which cplex+ finds solutions in the given timeout whereas cplex- does not.
There is no such example for the opposite case.
Overall, the inclusion of WSPT does not bring substantial improvements, and the rules can be ignored rather safely.

For the CP case, the situation looks quite different.
In fact, we can observe that the WSPT rules even \emph{harm} the process of finding solutions.
There is a whole series in which adding the WSPT rule yields to no solution being found.
The maximum gaps indicates that this happens quite frequently even for small instances.
Hence, it seems more advisable to renounce this additional knowledge during search or incorporate it in another way.

\subsection{Analysis of Ontime-Results}
\begin{figure}[t]
    \centering
    \includegraphics[width=\textwidth]{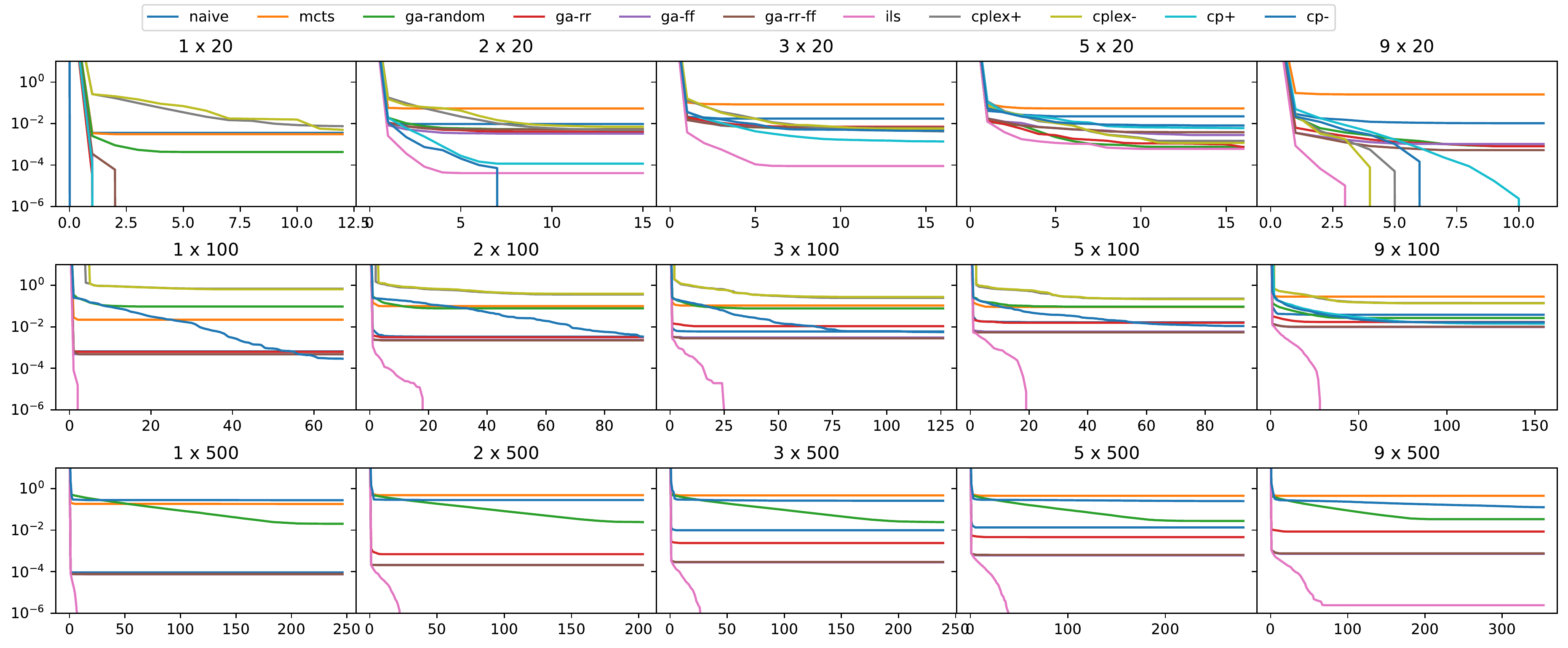}
    \caption{Average Performances of the Algorithms for $d$ index $40\%$, averaged.}
    \label{fig:results:timelines}
\end{figure}
In order to give not only insights into the final results but also into the runtime behavior of the algorithms, Fig. \ref{fig:results:timelines} shows gaps  of the best observed solution until some predefined runtime for each algorithm (on average).
That is, for each problem instance, we consider the best \emph{final} result observed by any algorithm and then compute, for any intermediate point of time and any algorithm, the gap between the currently best seen solution of the respective algorithm, and the finally best solution.
These gaps are then averaged for each point of time and merged into a mean gap curve for each algorithm.
To exclude secondary effects of infinity for cases where no solution is found, we consider the 20\% trimmed mean.
On the time axis, which shows the passed time in seconds, the charts are cut at the point of time where none of the algorithms will make any more progress, so the time axis does most often not exploit the full timeout of the respective problem setup.
To differentiate on a fine level, plots are depicted on a log-scale so that the best solution eventually drops to minus infinity, which nicely distinguishes the (avg.) point in time in which the algorithm hits the eventual optimum.

In the given charts, we can observe a remarkable dominance of ILS in time.
In fact, ILS not only provides the best overall results but also is even dominant with respect to performance per time.
That is, in the large majority of the experiments, at each point of time during the run of ILS, the best observed performance up to that time is at least as good and often better than the best performance observed by any other algorithm to the same time.
In other words, ILS does not only find the best solutions but is even fastest in finding them, so the quality of ILS does not come at the cost of high runtimes.

Another remarkable observation is that the algorithms stall very early and that up to 99\% of the time elapse in vain without any further improvements.
In the most extreme case of 9 machines and 500 jobs, none of the algorithms can improve upon the best solution found within 200s, effectively yielding 5h of useless computation time for all of the algorithms.
It is particularly interesting that the algorithms are not leveled out at this point, which implies that the algorithms are somewhat incapable of leaving their local optima.
This is one explanation of the success of ILS, which is the only (local) algorithm with an unconditional and rather long traversal through the search space.
These insights suggest the potential of improvements by algorithms that are more exploratory in non-neighborhood regions.

\section{Conclusion}
In this paper, we have introduced the weighted flowtime minimization problem \probGraham in which the release dates of jobs are decision variables, which  are constrained by a global deadline for the latest possible release date.
To our knowledge, this is surprisingly the first paper to study this practically important problem.
We have shown the NP-completeness of the problem even for the case of only one machine and provided an empirical problem analysis that shows that naive solutions that pretend that no deadline exists often already exhibit quite good performances.
In an exhaustive study, we have shown that an iterative local search generally finds in the frame of some few minutes the best solutions that are found by an ensemble of 11 algorithms within a time frame of up to 5 hours, and that a naive solution that ignores the deadline most often achieves solutions that are only slightly worse than the best found solution.

We see several directions for future work: the most obvious extensions of this work imply the derivative of a lower bound for partial solutions, which can then be used to devise a B\&B algorithm, and a further analysis of exploratory algorithms similar to ILS that allow to better escape from local optima.
Another interesting next step is to consider related \emph{problems}, which make the setting more realistic by introducing stochastic aspects.
More precisely, we plan to study the problem in which only the \emph{expected} arrival times can be decided and that each job is associated with a distribution for the true arrival time around this mean.
Such a probabilistic model allows to cater for traffic jams in the supply and other external factors, which we expect to make the problem much more complicated and maybe require simulation.


\printbibliography
\end{document}